\documentclass[11pt,peerreview,draftclsnofoot,onecolumn,twosides]{IEEEtran}

\usepackage[cmex10]{amsmath}
\usepackage{epsfig}
\usepackage{amsthm}
\usepackage{url}
\usepackage{multirow}
\usepackage{rotating}
\usepackage{amssymb}
\usepackage{graphicx,amsfonts}
\usepackage{algorithm,algorithmic}

\usepackage{cite}
\ifCLASSINFOpdf
\else
\fi

\usepackage{stfloats}

\newcommand{\cH}{\mathcal{H}}

\newcommand{\cL}{\mathcal{L}}

\newcommand{\cC}{\mathcal{C}}

\newcommand{\bW}{\boldsymbol{W}}
\newcommand{\bZ}{\boldsymbol{Z}}

\newcommand{\bR}{\boldsymbol{R}}
\newcommand{\bS}{\boldsymbol{S}}

\newcommand{\bT}{\boldsymbol{T}}

\newcommand{\bPhi}{\boldsymbol{\Phi}}

\newcommand{\bx}{\boldsymbol{x}}
\newcommand{\bh}{\boldsymbol{h}}
\newcommand{\by}{\boldsymbol{y}}

\newcommand{\bz}{\boldsymbol{z}}
\newcommand{\ba}{\boldsymbol{a}}
\newcommand{\bb}{\boldsymbol{b}}

\newcommand{\bc}{\boldsymbol{c}}

\newcommand{\bw}{\boldsymbol{w}}

\newcommand{\bbf}{\boldsymbol{f}}
\newcommand{\bg}{\boldsymbol{g}}
\newcommand{\bZero}{\boldsymbol{0}}

\newcommand{\beq}{\begin{equation}}
\newcommand{\eeq}{\end{equation}}
\newcommand{\beqn}{\begin{eqnarray}}
\newcommand{\eeqn}{\end{eqnarray}}
\newcommand{\beqns}{\begin{eqnarray*}}
\newcommand{\eeqns}{\end{eqnarray*}}

\newcommand{\R}{\mathbb{R}}
\newcommand{\HH}{\mathbb{H}}
\newcommand{\XX}{\mathbb{X}}
\newcommand{\C}{\mathbb{C}}
\newcommand{\A}{\mathbb{A}}

\newcommand{\F}{\mathbb{F}}
\newcommand{\N}{\mathbb{N}}

\newcommand{\frechet}{\textrm{Fr\'{e}chet }}
\newcommand{\fredif}{\textrm{Fr\'{e}chet differentiable }}

\makeatletter
\newcommand{\bdiv}{\mathop{\operator@font div}}
\makeatother

\makeatletter
\newcommand{\diag}{\mathop{\operator@font diag}}
\makeatother

\makeatletter
\newcommand{\conv}{\mathop{\operator@font conv}}
\makeatother

\makeatletter
\newcommand{\sign}{\mathop{\operator@font sign}}
\makeatother \makeatletter

\makeatletter
\newcommand{\proj}{\mathop{\operator@font proj}}
\makeatother \makeatletter

\makeatletter
\newcommand{\spa}{\mathop{\operator@font span}}
\makeatother \makeatletter

\makeatletter
\newcommand{\epi}{\mathop{\operator@font epi}}
\makeatother \makeatletter

\makeatletter
\newcommand{\dom}{\mathop{\operator@font dom}}
\makeatother \makeatletter

\newtheorem{thm}{Theorem}[section]
\newtheorem{prop}{Proposition}[section]

\newtheorem{lem}[thm]{Lemma}

\theoremstyle{remark}

\theoremstyle{definition}

\theoremstyle{definition}

\begin{document}
\title{Extension of Wirtinger's Calculus to Reproducing Kernel Hilbert Spaces and the Complex Kernel LMS}

\author{Pantelis Bouboulis,~\IEEEmembership{Member,~IEEE,}
        and~Sergios Theodoridis,~\IEEEmembership{Fellow,~IEEE}% <-this % stops a space
\thanks{Copyright (c) 2010 IEEE. Personal use of this material is permitted.
However, permission to use this material for any other purposes must be obtained from the
IEEE by sending a request to pubs-permissions@ieee.org.}
\thanks{P. Bouboulis is with the Department
of Informatics and Telecommunications, University of Athens, Greece,
e-mail: bouboulis@di.uoa.gr.}% <-this % stops a space
%\thanks{Manuscript received ....}
\thanks{S. Theodoridis is with the Department
of Informatics and Telecommunications, University of Athens, Greece,
and the Research Academic Computer Technology Institute, Patra, Greece.
e-mail: stheodor@di.uoa.gr.}
}


\markboth{IEEE Transactions on Signal Processing}%
{Bouboulis \MakeLowercase{\textit{et al.}}: Wirtinger Calculus and Complex Kernel LMS}



\maketitle

\begin{abstract}
%\boldmath
Over the last decade, kernel methods for nonlinear processing have successfully been used  in the machine learning community. The primary mathematical tool employed in these methods is the notion of the Reproducing Kernel Hilbert Space. However, so far, the emphasis  has been on batch techniques. It is only recently, that online techniques have been considered in the context of adaptive signal processing tasks. Moreover, these efforts have only been focussed on real valued data sequences.  To the best of our knowledge, no adaptive kernel-based strategy has been developed, so far, for complex valued signals. Furthermore, although the real reproducing kernels are used in an increasing number of machine learning problems, complex kernels have not, yet, been used, in spite of their potential interest in applications that deal with complex signals, with Communications being a typical example. In this paper, we present a general framework to attack the problem of adaptive filtering of complex signals, using either real reproducing kernels, taking advantage of a technique called \textit{complexification} of real RKHSs,  or complex reproducing kernels, highlighting the use of the complex gaussian kernel.

In order to derive gradients of operators that need to be defined on the associated complex RKHSs, we employ the powerful tool of Wirtinger's Calculus, which has recently attracted attention in the signal processing community. Wirtinger's calculus simplifies computations and offers an elegant tool for treating complex signals. To this end, in this paper, the notion of Wirtinger's calculus is extended, for the first time, to include complex RKHSs and use it to derive several realizations of the Complex Kernel Least-Mean-Square (CKLMS) algorithm. Experiments verify that the CKLMS offers significant performance improvements over several linear and nonlinear algorithms, when dealing with nonlinearities.
\end{abstract}

%\begin{IEEEkeywords}
%Complex RKHS, Wirtinger's Calculus, KLMS, Adaptive filter, complex signal processing
%\end{IEEEkeywords}

% \fi
\IEEEpeerreviewmaketitle

%--------------------------------------------------------------------------------
\section{Introduction}\label{SEC:Intro}
%--------------------------------------------------------------------------------

\IEEEPARstart{P}{}rocessing in Reproducing Kernel Hilbert Spaces (RKHSs), in the context of online learning, is gaining in popularity within the Machine Learning and Signal Processing communities \cite{LiPokPrin, KivSmoWil, EngManMe, SlaTheYam, SlaTheYam2, SlaThe}. The main advantage of  mobilizing the tool of RKHSs is that the original nonlinear task is ``transformed" into a linear one, which can be solved by employing an easier ``algebra".  Moreover, different types of nonlinearities can be treated in a unifying way, with no effect on the mathematical derivation of the algorithms, except at the final implementation stage. The main concepts of this procedure can be summarized in the following two steps: 1) Map the finite dimensionality input data from the input space $F$ (usually $F\subset \R^\nu$) into a higher dimensionality (possibly infinite) RKHS $\cH$ and 2) Perform a linear processing (e.g., adaptive filtering) on the mapped data in $\cH$. The procedure is equivalent with a non-linear processing (non-linear filtering) in $F$.

An alternative way of describing this process is through the popular \textit{kernel trick} \cite{SchoSmo, TheoKou}: Given an algorithm, which can be formulated in terms of dot products, one can construct an alternative algorithm by replacing each one of the dot products with a positive definite kernel $\kappa$.  The specific choice of kernel implicitly defines  a RKHS with an appropriate inner product. Furthermore, the choice of kernel also defines the type of nonlinearity that underlies the model to be used. The main representatives of this class of algorithms are the celebrated \textit{support vector machines} (SVMs), which have dominated the research in machine learning over the last decade \cite{TayChr}. Besides SVMs and the more recent applications in adaptive filtering, there is a plethora of other scientific domains that have gained from adopting kernel methods (e.g., image processing and denoising \cite{KimFraScho, BouSlaThe}, principal component analysis \cite{SchoSmoMu}, clustering \cite{FilCaMaRo}, e.t.c.).

In classification tasks (which have been the dominant applications of kernel methods) the use of complex reproducing kernels is meaningless, since no arrangement can be derived in complex domains and the necessary separating hypersurfaces cannot be defined. Consequently, all known kernel based applications, as they emerged from the specific background, use real-valued kernels and they are able to deal with real valued data sequences only.  To our knowledge, no kernel-based strategy has been developed, so far, that is able to effectively deal with complex valued signals.

In this paper, we present a general framework to address the problem of adaptive filtering of complex signals, using either real reproducing kernels, taking advantage of a technique called \textit{complexification} of real RKHSs,  or complex reproducing kernels, highlighting mostly the use of the complex gaussian kernel. Although the real gaussian RBF kernel has become quite popular and it has been used in many applications, the complex gaussian RBF kernel, while known to the mathematicians (especially those working on Reproducing Kernel Hilbert Spaces or Functional Analysis), it has rather remained in obscurity in the Machine Learning and Signal Processing communities. Even though the presented framework has a broad range and may be applied to generalize a wide variety of kernel methods to the complex domain, this work focuses on the recently developed Kernel LMS (KLMS) \cite{LiPokPrin}, \cite{LiuPriHay}.

To compute the gradients of cost functions that are defined on the complex RKHSs, the principles of Wirtinger's calculus are employed. Wirtinger's calculus \cite{Wirti} has recently attracted attention in the signal processing community, mainly in the context of complex adaptive filtering \cite{Picin95, ManGoh, Adali10, Adali08a, Adali08b, LiAda08, LiAda09, AdaHay}, as a means of computing, in an elegant way,  gradients of real valued cost functions defined on complex domains ($\C^\nu$). To this end, the main ideas and theorems of Wirtinger's calculus are generalized to general complex Hilbert spaces for the first time.

To summarize, the main contributions of this paper are:  a) the development of a wide framework that allows real-valued kernel algorithms to be extended to treat complex data effectively, taking advantage of a technique called \textit{complexification} of real RKHSs, b) to elevate from obscurity the complex Gaussian kernel as a tool for kernel based adaptive processing of complex signals, c) the extension of \textit{Wirtinger's Calculus} in complex RKHSs as a means for an elegant and efficient computation of the gradients, which are involved in the derivation of adaptive learning algorithms, and d) the development of several realizations of the Complex Kernel LMS (CKLMS) algorithm, by exploiting the extension of Wirtinger's calculus and the generated  complex RKHSs.

The paper is organized as follows. We start with an introduction to RKHSs in Section \ref{SEC:PRELIM}, which includes real and complex kernels, before we briefly review the KLMS algorithm in Section \ref{SEC:KLMS}. In Section \ref{SEC:Complexification}, we describe the complexification procedure of a real RKHS, that provides a framework to develop complex kernel methods, based on popular real valued reproducing kernels (e.g., gaussian, polynomial, e.t.c.). A brief introduction on Wirtinger's Calculus in finite dimensional spaces can be found in Section \ref{SEC:Wirtinger}. The main notions of the extended Wirtinger's Calculus on general Hilbert spaces are summarized in Section \ref{SEC:hilbert_wirti} and the CKLMS is developed thereafter in Section \ref{SEC:CKLMS}. Finally, experimental results and conclusions are provided in Sections \ref{SEC:Experim} and \ref{SEC:Concl}. Throughout the paper, we will denote the set of all integers, real and complex numbers by $\N$, $\R$ and $\C$ respectively. Vector or matrix valued quantities appear in boldfaced symbols.

%--------------------------------------------------------------------------------
\section{Reproducing Kernel Hilbert Spaces}\label{SEC:PRELIM}
%--------------------------------------------------------------------------------
In this section, we briefly describe the theory of Reproducing Kernel Hilbert Spaces. Since we are interested on both real and complex kernels, we recall the basic facts on RKHS associated with a general field $\F$, which can be either $\R$ or $\C$. However, we highlight the basic differences between the two cases. The material presented here may be found with more details in \cite{Saitoh} and \cite{Paulsen}.

%--------------------------------------------------------------------------------
\subsection{Basic Definitions}\label{SEC:Ba_def}
Given a function $\kappa:X\times X\rightarrow\F$ and $x_1,\dots,x_N \in X$, the matrix\footnote{The term $(K_{i,j})^N$ denotes a square $N\times N$ matrix.} $K=(K_{i,j})^N$ with elements $K_{i,j}=\kappa(x_i,x_j)$, for $i,j=1,\dots,N$, is called the \textit{Gram matrix} (or \textit{kernel matrix}) of $\kappa$ with respect to $x_1,\dots,x_N$.
A Hermitian matrix $K=(K_{i,j})^N$ satisfying
\begin{align*}
c^H\cdot K\cdot c=\sum_{i=1,j=1}^{N,N} c_i^* c_j K_{i,j}\geq 0,
 \end{align*}
for all $c_i\in\F$, $i=1,\dots,N$, where the notation $^*$ denotes the conjugate element, is called \textit{Positive Definite}.
In matrix analysis literature, this is the definition of a positive semidefinite matrix. However, since this is a rather cumbersome term and the distinction between positive definite and positive semidefinite matrices is not important in this paper, we employ the term positive definite in the way presented here. Furthermore, the term positive definite was introduced for the first time by Mercer in the kernel context (see \cite{Mercer2}). Let $X$ be a nonempty set. Then a function $\kappa:X\times X\rightarrow\F$, which for all $N\in\N$ and all $x_1,\dots,x_N\in X$ gives rise to a positive definite Gram matrix $K$, is called a \textit{Positive Definite Kernel}.
In the following, we will frequently refer to a positive definite kernel simply as \textit{kernel}.

Next, consider a linear class $\cH$ of complex valued functions $f$ defined on a set
$X$. Suppose further, that in $\cH$ we can define an inner product
$\langle\cdot,\cdot\rangle_\cH$ with corresponding norm
$\|\cdot\|_\cH$ and that $\cH$ is complete with respect to that
norm, i.e., $\cH$ is a Hilbert space. We call $\cH$ a
\textit{Reproducing Kernel Hilbert Space (RKHS)}, if for all $y\in
X$ the evaluation functional $T_y:\cH\rightarrow\F:\;T_y(f)=f(y)$ is
a linear continuous (or, equivalently, bounded) operator. If this is true,
then by the Riesz's representation theorem, for all $y\in X$ there
is a function $g_y\in\cH$ such that $T_y(f)=f(y)=\langle f,
g_y\rangle_\cH$. The function $\kappa:X\times
X\rightarrow\F:\;\kappa(x,y)=g_y(x)$ is called a \textit{reproducing
kernel} of $\cH$. It can be easily proved that the function $\kappa$
is a positive definite kernel.

Alternatively, we can define a RKHS as a Hilbert space $\cH$ for which there exists a function $\kappa:X\times X\rightarrow\F$ with the following two important properties:
\begin{enumerate}
\item For every $x\in X$, $\kappa(\cdot,x)$ belongs to $\cH$.
\item $\kappa$ has the so called \textit{reproducing property}, i.e.,
\begin{align}\label{EQ:rep_prop}
f(x)=\langle f,\kappa(\cdot, x)\rangle_\cH, \textrm{ for all } f\in\cH,
\end{align}
in particular $\kappa(x,y)=\langle \kappa(\cdot, y), \kappa(\cdot, x)\rangle_\cH$.
\end{enumerate}

It has been shown (see \cite{Aron50}) that to every positive definite kernel $\kappa$ there corresponds one and only one class of functions $\cH$ with a uniquely determined inner product in it, forming a Hilbert space and admitting $\kappa$ as a reproducing kernel. In fact, the kernel $\kappa$ produces the entire space $\cH$, i.e.,
$\cH=\overline{\spa\{\kappa(x,\cdot)|x\in X\}}$\footnote{The overbar denotes the closure of the set.}.
The map $\Phi:X\rightarrow\cH:\Phi(x)=\kappa(\cdot,x)$ is called the \textit{feature map} of $\cH$.  Recall, that in the case of complex Hilbert spaces (i.e., $\F=\C$) the inner product is sesqui-linear (i.e., linear in one argument and antilinear in the other)  and Hermitian:
\begin{align*}
\langle a f + b g, h\rangle_{\cH} &= a\langle  f, h\rangle_{\cH} + b \langle g, h\rangle_{\cH},\\
\langle f , a g + b h\rangle_{\cH} &= a^*\langle  f, g\rangle_{\cH} + b^* \langle f, h\rangle_{\cH},\\
\langle f, g\rangle_{\cH}^* &= \langle  g, f\rangle_{\cH},
\end{align*}
for all $f,g,h \in\cH$, and $a,b\in\C$.
In the real case, the condition $\kappa(x,y)=\langle \kappa(\cdot, y), \kappa(\cdot, x)\rangle_\cH$ may be replaced by $\kappa(x,y)=\langle \kappa(\cdot, x), \kappa(\cdot, y)\rangle_\cH$. However, since in the complex case the inner product is Hermitian, the aforementioned condition is equivalent to $\kappa(x,y)=\left(\langle \kappa(\cdot, x), \kappa(\cdot, y)\rangle_\cH\right)^*$.
One of the most important properties of RKHSs is that norm convergence implies pointwise convergence. More precisely, let $\{f_n\}_{n\in\N}\subset\cH$ be a sequence such that $\lim_{n}\|f_n - f\|=0$, for some $f\in\cH$. Then, the continuity of $T_x$ gives $\lim_{n} f_n(x) = \lim_{n} T_x(f_n) = T_x(f) = f(x)$,
for all $x\in X$.

Although, the underlying theory has been developed by the mathematicians for  general complex reproducing kernels and their associated RKHSs, only the real kernels have been considered by the machine learning community. One of the most widely used kernel is the \textit{Gaussian RBF}, i.e.,
\begin{align}\label{EQ:real_gaussian_kernel}
\kappa_{\sigma,\R^d}(\bx,\by) : = \exp\left(-\frac{\sum_{i=1}^{d}(x_i-y_i)^2}{\sigma^2}\right),
\end{align}
defined for $\bx, \by \in \R^d$, where $\sigma$ is a free positive parameter.  Another popular kernel is the \textit{polynomial kernel}: $\kappa_d(\bx,\by) : = \left(1 + \bx^T \by\right)^d$,
for $d\in\N$. Many more can be found in the related literature \cite{SchoSmo, TheoKou, TayChr}.

Complex reproducing kernels, that have been extensively studied by the mathematicians, are, among others, the \textit{Szego kernels}, i.e, $\kappa(z,w)=\frac{1}{1-w^*z}$, for Hardy spaces on the unit disk, and the Bergman kernels, i.e., $\kappa(z,w) = \frac{1}{(1 - w^* z)^2}$, for Bergman spaces on the unit disk, where $|z|, |w| < 1$ \cite{Paulsen}. In the following, we discuss another complex kernel that has remained relatively unknown in the Machine Learning and Signal Processing societies.

%--------------------------------------------------------------------------------
\subsection{The Complex Gaussian Kernel}\label{coml_gaus}
Consider the complex valued function
\begin{align}\label{EQ:complex_gaussian_kernel}
\kappa_{\sigma,\C^d}(\bz,\bw) : = \exp\left(-\frac{\sum_{i=1}^{d}(z_i-w_i^*)^2}{\sigma^2}\right),
\end{align}
defined on $\C^d\times\C^d$, where $\bz,\bw\in\C^d$, $z_i$ denotes the $i$-th component of the complex vector $\bz\in\C^d$ and $\exp$ is the extended exponential function in the complex domain. It can be shown that $\kappa_{\sigma,\C^d}$ is a complex valued kernel, which we call the \textit{complex Gaussian kernel} with parameter $\sigma$. Its restriction $\kappa_{\sigma}:=\left(\kappa_{\sigma,\C^d}\right)_{|\R^d\times\R^d}$ is the well known \textit{real Gaussian kernel}.
An explicit description of the RKHSs of these kernels, together with some important properties can be found in \cite{SteinHuSco}.

%--------------------------------------------------------------------------------
\section{Kernel Least Mean Square Algorithm}\label{SEC:KLMS}
%--------------------------------------------------------------------------------

In a typical LMS filter the goal is to learn a linear input-output mapping $f:X\rightarrow\R:f(\bx)=\bw^T\bx$, $X\subset\R^\nu$, based on a sequence of examples $(\bx(1),d(1)), (\bx(2),d(2)), \dots, (\bx(N),d(N))$, so that to minimize the mean square error, $E\left[|d(n) - \bw^T\bx(n)|^2\right]$. To this end, the gradient descent rationale is employed and at each time instant,  $n=1,2,\dots,N$, the gradient of the mean square error, i.e., $-2 E[e(n)\bx(n)]$, is estimated via its current measurement, i.e.,  $\hat E[e(n)\bx(n)]=e(n)\bx(n)$, where $e(n) = d(n) - \bw(n-1)^T\bx(n)$ is the a-priori error at instance $n=2,\dots,N$. It takes a few lines of elementary algebra to deduce that the update of the unknown vector parameter is: $\bw(n) = \bw(n-1) + \mu e(n) \bx(n)$, where $\mu$ is the parameter controlling the step update. If we take the initial value of $\bw$ as $\bw(0)=\bZero$, then the repeated application of the update equation yields:
\begin{align}\label{EQ:LMS1}
\bw(n) = \mu \sum_{k=1}^n e(k) \bx(k)
\end{align}
Hence, for the filter output at instance $n$ we have:
\begin{align}\label{EQ:LMS2}
\hat d(n) = \bw(n-1)^T\bx(n) = \mu \sum_{k=1}^{n-1} e(k) \bx(k)^T \bx(n),
\end{align}
for $n=1,2,\dots,N$. Equation (\ref{EQ:LMS2}) is expressed in terms of inner products only, hence it allows for the application of the kernel trick. Thus, the filter output of the KLMS at instance $n$ is
\begin{align}\label{EQ:KLMS2}
\hat d(n) =  \left\langle \bx(n), \bw(n-1)\right\rangle_{\cH} = \mu\sum_{k=1}^{n-1} e(k) \kappa\left( \bx(n), \bx(k)\right),
\end{align}
\begin{align}\label{EQ:KLMS1}
\textrm{while }\hspace{4em} \bw(n) = \mu \sum_{k=1}^n e(k) \kappa(\cdot, \bx(k)),
\end{align}
for $n=1,2,\dots,N$.

Another, more formal, way of developing the KLMS is the following. First, we transform the input space $X$ to a high dimensional feature space $\cH$, through the (implicit) mapping $\Phi:X\rightarrow\cH$, $\Phi(\bx)=\kappa(\cdot,\bx)$. Thus, the training examples become
$(\Phi(\bx(1)), d(1)), \dots, (\Phi(\bx(N)), d(N)).$
We apply the LMS procedure on the transformed data, with the linear filter output $\hat d(n)=\langle \Phi(\bx(n)), \bw\rangle_{\cH}$. The model $\langle \Phi(\bx), \bw\rangle_{\cH}$ is more representative than the simple $\bw^T\bx$, since it includes the nonlinear modeling through the presence of the kernel. The objective now becomes to minimize the cost function
$E\left[|d(n) - \langle \Phi(\bx(n)), \bw\rangle_{\cH}|^2\right]$
(see \cite{Luen84}). Using the notion of the \frechet derivative \cite{Luen84, LiuSob, Balak}, which has to be mobilized, since the dimensionality of the RKHS may be infinite, we are able to derive the gradient of the aforementioned cost function with respect to $\bw$, if we estimate it by its current measurement $|d(n) - \langle \Phi(\bx(n)), \bw\rangle|^2$.  Thus the respective gradient is $-2e(n) \Phi(\bx(n))$. It has to be emphasized, that now $\bw$ is not a vector, but a function, i.e., a point in the linear Hilbert space. It turns out that the update of the KLMS is given by $\bw(n)=\bw(n-1) + \mu e(n) \Phi(\bx(n))$, where $e(n)=d(n)-\hat d(n)$. From this update, following the same procedure as in LMS and applying the reproducing property, we obtain equations (\ref{EQ:KLMS2}) and (\ref{EQ:KLMS1}), which are at the core of the KLMS algorithm. More details and the algorithmic implementation may be found in \cite{LiuPriHay}.

Note that in a number of attempts to kernelize known algorithms, that are cast in inner products, the kernel trick is, usually, used in a "black box" rationale, without consideration of the problem in the RKH space, in which the (implicit) processing is carried out. Such an approach, often, does not allow for a deeper understanding of the problem, especially if a further theoretical analysis is required. Moreover, in our case, such a ``blind" application of the kernel trick on a standard complex LMS form, can only lead to spaces defined by complex kernels, as it will become clear soon. Complex RKH spaces, that are built around complexification of real  kernels, do not result as a direct application of the standard kernel trick.

%--------------------------------------------------------------------------------
\section{Complexification of real Reproducing Kernel Hilbert Spaces}\label{SEC:Complexification}
%--------------------------------------------------------------------------------
To generalize the kernel adaptive filtering algorithms on complex domains, we need a universal framework regarding complex RKHSs. A first straightforward approach is to use directly a complex RKHS, using one of the complex kernels given in section \ref{SEC:PRELIM}. In this section, we present an alternative simple technique called \textit{complexification} of real RKHSs, which has the advantage of allowing  modeling in complex RKHSs using popular well-established and well understood, from a performance point of view, real kernels (e.g., gaussian, polynomial, e.t.c.).

Let $X\subseteq\R^\nu$. Define $X^2\equiv X\times X\subseteq\R^{2\nu}$ and $\XX=\{\bx+i\by, \bx,\by\in X\}\subseteq\C^{\nu}$ equipped with a complex product structure. Let $\cH$ be a real RKHS associated with a real kernel $\kappa$ defined on $X^2\times X^2$ and let $\langle\cdot,\cdot\rangle_\cH$ be its corresponding inner product. Then, every $f\in\cH$ can be regarded as a function defined on either $X^2$ or $\XX$, i.e., $f(\bz) = f(\bx+i\by) = f(\bx,\by)$.

Next, we define $\cH^2=\cH\times\cH$. It is easy to verify that $\cH^2$ is also a Hilbert Space with inner product
\begin{align}
\langle \bbf, \bg\rangle_{\cH^2} = \langle f_1, g_1\rangle_\cH + \langle f_2, g_2\rangle_\cH,
\end{align}
for $\bbf=(f_1,f_2)$, $\bg=(g_1,g_2)$. Our objective is to enrich $\cH^2$ with a complex structure. We address this problem using the complexification of the real RKHS  $\cH$. To this end, we define the space $\HH= \{\bbf=f_1 + i f_2;\;f_1,f_2\in\cH\}$
equipped with the complex inner product:
\begin{align*}
\langle \bbf, \bg\rangle_{\HH}= \langle f_1, g_1\rangle_\cH + \langle f_2, g_2\rangle_\cH +
                 i\left(\langle f_2, g_1\rangle_\cH - \langle f_1, g_2\rangle_\cH\right),
\end{align*}
for $\bbf=f_1 + if_2$, $\bg=g_1 + i g_2$. Hence, $\bbf, \bg:\XX\subseteq\C^{\nu}\rightarrow \C$. It is not difficult to verify that $\HH$ is a complex RKHS with kernel $\kappa$ \cite{Paulsen}. We call $\HH$ the complexification of $\cH$. It can readily be seen, that, although $\HH$ is a complex RKHS, its respective kernel is real (i.e., its imaginary part is equal to zero).

To complete the presentation of the required framework for working on complex RKHSs using this rationale, we need a technique to implicitly map the samples data from the complex input space to the complexified RKHS $\HH$. This can be done using the simple rule:
\begin{align}\label{EQ:Phi_map}
\bPhi(\bz)=\bPhi(\bx+ i\by) = \bPhi(\bx,\by) = \Phi(\bx,\by) + i\Phi(\bx,\by),
\end{align}
where $\Phi$ is the feature map of the real reproducing kernel $\kappa$, i.e., $\Phi(\bx,\by)=\kappa(\cdot, (\bx, \by))$. It must be emphasized, that $\bPhi$ is not the feature map associated with the complex RKHS $\HH$. Furthermore, the employed kernel is a real one.  Therefore, the algorithms derived using this approach cannot be reproduced, if one blindly applies the kernel trick using any complex kernel. However, observe that:
\begin{align*}
\langle\bPhi(\bz), \bPhi(\bz')\rangle_{\HH} = 2\langle\Phi(\bx,\by), \Phi(\bx',\by')\rangle_{\cH}\\
 = 2\kappa( (\bx,\by), (\bx',\by') ).
\end{align*}
This relation implies that the complexification procedure is equivalent with the following \textit{complexified real kernel trick}: Given an algorithm, which is formulated in terms of complex dot products (i.e, $\bw^H\bz$, where $\bz=\bx + i\by$, $\bw=\bw_1 + i\bw_2$), one can construct an alternative algorithm by replacing each one of the complex dot products with a positive definite \textit{real kernel} $\kappa$, with arguments the extended real vectors of $\bz$ and $\bw$ (i.e., $\kappa( (\bx, \by), (\bw_2, \bw_2) )$).

%--------------------------------------------------------------------------------
\section{Wirtinger's Calculus on $\C$}\label{SEC:Wirtinger}
%--------------------------------------------------------------------------------
Wirtinger's calculus \cite{Wirti} is enjoying increasing popularity in the signal processing community mainly in the context of complex adaptive filtering \cite{Picin95, ManGoh, Adali10, Adali08a, Adali08b, LiAda08, LiAda09, AdaHay}, as a means to compute, in an elegant way,  gradients of real valued cost functions that are defined on complex domains ($\C^\nu$). The Cauchy-Riemann conditions dictate that such functions are not holomorphic (except from the case where the function is a constant) and therefore the complex derivative cannot be used. Instead, if we consider that the cost function is defined on a Euclidean domain with a double dimensionality ($\R^{2\nu}$), then the real derivatives may be employed. The price of this approach is that the computations may become cumbersome and tedious. Wirtinger's calculus provides an alternative equivalent formulation, that is based on simple rules and principles and which bears a great resemblance to the rules of the standard complex derivative. In this section, we present the main notions of Wirtinger's calculus for functions defined on complex domains. These ideas are, subsequently, extended in section \ref{SEC:hilbert_wirti} to include the case of general complex Hilbert spaces.

Let  $f:\C\rightarrow\C$ be a complex function defined on $\C$. Obviously, such a function may be regarded as either defined on $\R^2$ or $\C$ (i.e., $f(z)=f(x+iy)=f(x,y)$). Furthermore, it may be regarded as either a complex valued function, $f(x,y)=u(x,y)+iv(x,y)$ or as a vector valued function $f(x,y)=(u(x,y), v(x,y))$. We will say that $f$ is \textit{differentiable in the real sense} if $u$ and $v$ are differentiable.  It turns out that, when the complex structure is considered, the real derivatives may be described using an equivalent and more elegant formulation, which bears a surprising resemblance with the complex derivative. In fact, if the function $f$ is \textit{differentiable in the complex sense} (i.e. the complex derivative exists), the developed derivatives coincide with the complex ones. Although this methodology is known for some time in the German speaking countries and it has been applied to practical applications \cite{Brandwood, VanDeBos}, only recently has attracted the attention of the signal processing community, mostly in the context of works that followed Picinbono's paper on widely linear estimation filters \cite{Picin95}.

The \textit{Wirtinger's derivative} (or \textit{W-derivative} for short) of $f$ at a point $c$ is defined as follows
\begin{align}\label{EQ:wirti_der}
\frac{\partial f}{\partial z}(c) &= \frac{1}{2}\left(\frac{\partial f}{\partial x}(c)  -i \frac{\partial f}{\partial y}(c)\right)
 = \frac{1}{2}\left(\frac{\partial u}{\partial x}(c) + \frac{\partial v}{\partial y}(c)\right)
      + \frac{i}{2}\left(\frac{\partial v}{\partial x}(c) - \frac{\partial u}{\partial y}(c)\right).
\end{align}
The \textit{conjugate Wirtinger's derivative} (or \textit{CW-derivative} for short) of $f$ at $c$ is defined by:
\begin{align}\label{EQ:conj_wirti_der}
\frac{\partial f}{\partial z^*}(c) &= \frac{1}{2}\left(\frac{\partial f}{\partial x}(c)  +i \frac{\partial f}{\partial y}(c)\right)
 = \frac{1}{2}\left(\frac{\partial u}{\partial x}(c) - \frac{\partial v}{\partial y}(c)\right)
      + \frac{i}{2}\left(\frac{\partial v}{\partial x}(c) + \frac{\partial u}{\partial y}(c)\right).
\end{align}

The following properties can be proved \cite{Delga, Bou_Wirti, LiAda08}:
\begin{enumerate}
\item If $f$ has a Taylor series expansion with respect to $z$ (i.e., it is holomorphic) around $c$, then $\frac{\partial f}{\partial z^*}(c)=0$.
\item If $f$ has a Taylor series expansion with respect to $z^*$ around $c$, then $\frac{\partial f}{\partial z}(c)=0$.
\item $\left(\frac{\partial f}{\partial z}(c)\right)^* = \frac{\partial f^*}{\partial z^*}(c)$.
\item $\left(\frac{\partial f}{\partial z^*}(c)\right)^* = \frac{\partial f}{\partial z^*}(c)$.
\item Linearity: If $f,g$ are differentiable in the real sense at $c$ and $\alpha, \beta\in\C$, then
\begin{align*}
\frac{\partial (\alpha f + \beta g)}{\partial z}(c) = \alpha\frac{\partial f}{\partial z}(c) + \beta\frac{\partial g}{\partial z}(c),\quad
\frac{\partial (\alpha f + \beta g)}{\partial z^*}(c) = \alpha\frac{\partial f}{\partial z^*}(c) + \beta\frac{\partial g}{\partial z^*}(c)
\end{align*}
\item Product Rule: If $f$, $g$ are  differentiable in the real sense at $c$, then
\begin{align*}
\frac{\partial (f \cdot g)}{\partial z}(c) = \frac{\partial f}{\partial z}(c)g(c) + f(c)\frac{\partial g}{\partial z}(c),\quad
\frac{\partial (f \cdot g)}{\partial z^*}(c) = \frac{\partial f}{\partial z^*}(c)g(c) + f(c)\frac{\partial g}{\partial z^*}(c).
\end{align*}
\item Division Rule: If $f$, $g$ are  differentiable in the real sense at $c$ and $g(c)\not=0$, then
\begin{align*}
\frac{\partial(\frac{f}{g})}{\partial z}(c) = \frac{\frac{\partial f}{\partial z}(c) g(c) - f(c)\frac{\partial g}{\partial z}(c)}{g^2(c)},\quad
\frac{\partial(\frac{f}{g})}{\partial z^*}(c) = \frac{\frac{\partial f}{\partial z^*}(c) g(c) - f(c)\frac{\partial g}{\partial z^*}(c)}{g^2(c)}.
\end{align*}
\item Chain Rule: If $f$ is  differentiable in the real sense at $c$ and $g$ is differentiable in the real sense at $f(c)$, then
\begin{align*}
\frac{\partial g\circ f}{\partial z}(c) &= \frac{\partial g}{\partial z}(f(c))\frac{\partial f}{\partial z}(c) + \frac{\partial g}{\partial z^*}(f(c))\frac{\partial f^*}{\partial z}(c),\\
\frac{\partial g\circ f}{\partial z^*}(c) &= \frac{\partial g}{\partial z}(f(c))\frac{\partial f}{\partial z^*}(c) + \frac{\partial g}{\partial z^*}(f(c))\frac{\partial f^*}{\partial z^*}(c).
\end{align*}
\end{enumerate}

In view of the aforementioned properties, one might easily compute the W and CW derivatives of any complex function $f$, which is written in terms of $z$ and $z^*$, following the following simple tricks:

\textit{
\begin{quote}
\begin{itemize}
\item To compute the W-derivative of a function $f$, which is expressed in terms of $z$ and $z^*$, apply the usual differentiation rules considering $z^*$ as a constant.
\item To compute the CW-derivative of a function $f$, which is expressed in terms of $z$ and $z^*$, apply the usual differentiation rules considering $z$ as a constant.
\end{itemize}
\end{quote}
}

Note that any complex function $f(z)$, which is differentiable in the real sense, can be cast in terms of $z$ and $z^*$. For example, if the function $f(z)=f(x+iy)=f(x,y)$ is given in terms of $x$ and $y$, replacing $x$ by $(z+z^*)/2$ and $y$ by $(z-z^*)/2$ gives the result.  It should be emphasized, that these statements must be regarded as a simple computational trick rather than as a rigorous mathematical rule. This trick works well due to the aforementioned properties. Nonetheless, special care should be considered whenever these tricks are applied. For example, given the function $f(z)=|z|^2$, we might conclude that $\frac{\partial f}{\partial z^*}=0$, since if we consider $z$ as a constant, then $f(z)$ is also a constant. However, one might argue that since there isn't any rule regarding the complex norm, this rationale leads to an error. Undeniably, if one recasts $f$ as $f(z)=z z^*$, then one concludes that $\frac{\partial f}{\partial z^*}=z$ and $\frac{\partial f}{\partial z}=z^*$. Similar rules and principles hold for functions defined on $\C^\nu$ \cite{Delga}.

%--------------------------------------------------------------------------------
\section{Extension of Wirtinger's Calculus to general Hilbert spaces}\label{SEC:hilbert_wirti}
To apply minimization algorithms on real valued operators defined on complex RKHSs, we need to compute the associated gradients.
To this end, in this section, we generalize the main ideas and results of Wirtinger's calculus on general Hilbert spaces. We begin with a brief review of the \textit{\frechet derivative}, which generalizes differentiability to Hilbert spaces and which will be the basis for our discussion.

%---------------------------------------------------------------------------------------
\subsection{\frechet Derivatives}\label{EQ:frechet}
Since \frechet differentiability is not the mainstream of mathematical tools used in the Signal Processing and Machine Learning communities, we give here some basic definitions for the sake of clarity. Consider a Hilbert space $H$ over the field $F$  (typically $\R$ or $\C$). The operator $\bT:H\rightarrow F^\nu$ is said to be \textit{\fredif} at $f_0$, if there exists a linear continuous operator $\bW=(W_1,W_2,\dots,W_\nu):H\rightarrow\F^\nu$ such that
\begin{align}\label{EQ:frechet1}
\lim_{\|h\|_{H}\rightarrow 0}\frac{\left\|\bT(f_0+h)-\bT(f_0)- \bW(h)\right\|_{F^\nu}}{\|h\|_{H}}=0,
\end{align}
where $\|\cdot\|_H=\sqrt{\langle\cdot, \cdot\rangle_H}$ is the induced norm of the corresponding Hilbert Space. Note that $F^\nu$ is considered as a Banach space under the Euclidean norm.  The linear operator $\bW$ is called the \textit{\frechet derivative} and is usually denoted by $d\bT(f_0):H\rightarrow F^\nu$.  Observe that this definition is valid not only for Hilbert spaces, but for general Banach spaces too. However, since we are mainly interested at Hilbert spaces, we present the main ideas in this context. It can be proved that if such a linear continuous operator $\bW$ can be found, then it is unique (i.e., the derivative is unique) \cite{LiuSob}.
In the special case where $\nu=1$ (i.e., the operator $\bT$ takes values on $F$) using the \textit{Riesz's representation} theorem,  we may replace the linear continuous operator $\bW$ with an inner product. Therefore, the operator $T:H\rightarrow F$ is said to be \textit{\fredif} at $f_0$, iff there exists a $w\in H$, such that
\begin{align}\label{EQ:frechet2}
\lim_{\|h\|_{H}\rightarrow 0}\frac{T(f_0+h)-T(f_0)-\langle h, w\rangle_{H}}{\|h\|_{H}}=0,
\end{align}
where $\langle\cdot, \cdot\rangle_{H}$ is the dot product of the Hilbert space $H$ and $\|\cdot\|_H$ is the
induced norm. The element $w^*$ is usually called the gradient of $T$ at $f_0$ and it is denoted by $w^*=\nabla T(f_0)$.

For a general vector valued operator $\bT=(T_1,\dots,T_\nu):H\rightarrow F^\nu$, we may easily  derive that if $\bT$ is differentiable at $f_0$, then $T_\iota$ is differentiable at $f_0$, for all $\iota=1,2,\dots,\nu$, and that
\begin{align}\label{EQ:vector_grad}
d\bT(f_0)(h) = \left(\begin{matrix}\langle h, \nabla T_1(f_0)^*\rangle_{H}\cr
\vdots
\cr \langle h, \nabla T_\nu(f_0)^*\rangle_{H}\end{matrix}\right).
\end{align}
To prove this claim, consider that since $\bT$ is differentiable, there exists a continuous linear operator $\bW$ such that
\begin{align*}
\lim_{\|h\|_{H}\rightarrow 0}\frac{\left\|\bT(f_0+h)-\bT(f_0)- \bW(h)\right\|_{F^\nu}}{\|h\|_{H}}=0\Leftrightarrow\\
\lim_{\|h\|_{H}\rightarrow 0} \left( \sum_{\iota=1}^{\nu} \frac{\left|T_\iota(f_0+h)-T_\iota(f_0)- W_\iota(h)\right|_{F}^2}{\|h\|^2_{H}}  \right)  = 0,
\end{align*}
for all $\iota=1,\dots,\nu$. Thus,
\begin{align*}
\lim_{\|h\|_{H}\rightarrow 0} \left( \frac{T_\iota(f_0+h)-T_\iota(f_0)- W_\iota(h)}{\|h\|_{H}}  \right) = 0,
\end{align*}
for all $\iota=1,2,\nu$. The Riesz's representation theorem dictates that since $W_\iota$ is a continuous linear operator, there exists $w_\iota\in H$, such that $W_\iota(h)=\langle h, w_\iota\rangle_H$, for all $\iota=1,\dots,\nu$. This proves that $T_\iota$ is differentiable at $f_0$ and that $w_\iota^*=\nabla T_\iota(f_0)$, thus equation (\ref{EQ:vector_grad}) holds. The converse is proved similarly.

The notion of \textit{\frechet differentiability} may be extended to include also partial derivatives. Consider the operator $T:H^\mu\rightarrow F$ defined on the Hilbert space $H^\mu$ with corresponding inner product:
\begin{align*}
\langle \bbf, \bg\rangle_{H^\mu} = \sum_{\iota=1}^{\mu} \langle f_\iota, g_\iota\rangle_H,
\end{align*}
where $\bbf=(f_1,f_2,\dots f_\mu)$, $\bg=(g_1,g_2,\dots g_\mu)$. $T(\bbf)$ is said to be \textit{\fredif} at $\bbf_0$ with respect to $f_\iota$, iff there exists a $w\in H$, such that
\begin{align}\label{EQ:frechet3}
\lim_{\|h\|_{H}\rightarrow 0}\frac{T(\bbf_0 + [h]_\iota)-T(\bbf_0)-\langle [h]_\iota, w\rangle_{H}}{\|h\|_{H}}=0,
\end{align}
where $[h]_\iota=(0, 0, \dots, 0, h, 0, \dots, 0)$, is the element of $H^\mu$ with zero entries everywhere, except at place $\iota$.
The element $w^*$ is called the gradient of $T$ at $\bbf_0$ with respect to $f_\iota$ and it is denoted by $w^*=\nabla_\iota T(\bbf_0)$. The \frechet partial derivative at $\bbf_0$ with respect to $f_\iota$ is denoted by $\frac{\partial T}{\partial f_\iota}(\bbf_0)$, $\frac{\partial T}{\partial f_\iota}(\bbf_0)(h)=\langle [h]_{\iota}, w\rangle_{\HH}$.

Although it will not be used here, it is interesting to note, that it is also possible to define \frechet derivatives of higher order and a corresponding Taylor's series expansion. In this context, the $n$-th \frechet derivative of $\bT$ at $\bbf_0$, denoted as $d^{n}\bT(\bbf_0)$, is a multilinear\footnote{A function is called multilinear, if it is linear in each variable.} map. If $\bT$ has \frechet derivatives of any order, it can be expanded as a Taylor series \cite{Bong}, i.e.,
\begin{align}\label{EQ:frechet_Taylor}
\bT(\bbf_0+\bh) = \sum_{n=0}^{\infty} \frac{1}{n!}d^n\bT(\bbf_0)(\bh, \bh, \dots, \bh).
\end{align}
In relative literature the term $d^n\bT(\bc)(\bh, \bh, \dots, \bh)$ is often replaced by $d^n\bT(\bc)\cdot \bh^n$, which it denotes that the multilinear map $d^n\bT(\bc)$ is applied to $(\bh, \bh, \dots, \bh)$.

%---------------------------------------------------------------------------------------
\subsection{Complex Hilbert spaces}\label{SEC:Hilbert_complex}

Let $\cH$ be a real Hilbert space with inner product $\langle\cdot,\cdot\rangle_{\cH}$ and $\cH^2$, $\HH$ the Hilbert spaces defined as shown in section \ref{SEC:Complexification}.  In the following, the complex structure of $\HH$ will be used to derive derivatives similar to the ones obtained from Wirtinger's calculus on $\C$.

Consider the function $\bT:A\subseteq\HH\rightarrow\C$, $\bT(\bbf)=\bT(u_{\bbf}+iv_{\bbf})=T_r(u_{\bbf}, v_{\bbf}) + i T_i(u_{\bbf}, v_{\bbf})$, defined on an open subset $A$ of $\HH$, where $u_{\bbf}, v_{\bbf}\in \cH$ and $T_r, T_i$ are real valued functions defined on $\cH^2$. Any such function, $\bT$, may be regarded as defined either on a subset of $\HH$, or on a subset of $\cH^2$. Moreover, $\bT$ may be regarded either as a complex valued function, or as a vector valued function, which takes values in $\R^2$. Therefore, we may equivalently write:
\begin{align*}
\bT(\bbf) &= \bT(u_{\bbf}+iv_{\bbf}) = T_r(u_{\bbf}, v_{\bbf}) + iT_i(u_{\bbf}, v_{\bbf}),\\
\bT(\bbf)&= \left( T_r(u_{\bbf}, v_{\bbf}), T_i(u_{\bbf}, v_{\bbf}) \right).
\end{align*}
In the following, we will often change the notation according to the specific problem and consider any element of $\bbf\in\HH$ defined either as $\bbf=u_{\bbf} + i v_{\bbf}\in\HH$, or as $\bbf=(u_{\bbf}, v_{\bbf})\in \cH^2$. In a similar manner, any complex number may be regarded as either an element of $\C$, or as an element of $\R^2$.
We say that $\bT$ is \textit{\frechet complex differentiable} at $\bc\in\HH$, if there exists $\bw\in\HH$ such that:
\begin{align}\label{EQ:frechet_complex}
\lim_{\|\bh\|_{\HH}\rightarrow 0}\frac{\bT(\bc+\bh)-\bT(\bc)-\langle \bh, \bw\rangle_{\HH}}{\|\bh\|_{\HH}}=0.
\end{align}
Then $\bw^*$ is called the \textit{complex gradient} of $\bT$ at $\bc$ and it is denoted as $\bw^*\equiv\nabla \bT(\bc)$. The \frechet complex derivative of $\bT$ at $\bc$ is denoted as $d\bT(\bc)(\bh)=\langle \bh, \bw\rangle_{\HH}$.
This definition, although similar with the typical \frechet derivative, exploits the complex structure of $\HH$. More specifically, the complex inner product, that appears in the definition, forces a great deal of structure on $\bT$. Similarly to the case of ordinary complex functions, it is this simple fact that gives birth to all the important strong properties of the complex derivative. For example, it can be proved, that if $d\bT(\bc)$ exists, then so does $d^{n}\bT(\bc)$, for $n\in\N$.  If $\bT$ is differentiable at any $\bc\in \A$, $\bT$ is called \textit{\frechet holomorphic} in $A$, or \textit{\frechet complex analytic} in $A$, in the sense that it can be expanded as a Taylor series, i.e.,
\begin{align}\label{EQ:fre_complex_Taylor}
\bT(\bc+\bh) = \sum_{n=0}^{\infty} \frac{1}{n!}d^n\bT(\bc)(\bh, \bh, \dots, \bh).
\end{align}
The proof of this statement is out of the scope of this paper. The interested reader may dig deeper on this subject by referring to \cite{Bong}. We begin our study by exploring the relations between the complex \frechet derivative and the real \frechet derivatives. In the following, we will say that $\bT$ is \textit{\frechet differentiable in the complex sense}, if the complex derivative exists, and that $\bT$ is \textit{\frechet differentiable in the real sense}, if its real \frechet derivative exists (i.e., $\bT$ is regarded as a vector valued operator $\bT:\cH^2\rightarrow\cH^2$). Similarly, the expression ``\textit{$\bT$ is \frechet complex analytic at  $\bc$}'' means that $\bT$ is \frechet complex analytic at a neighborhood around $\bc$. We will say that $\bT$ is \textit{\frechet real analytic}, when both $T_r$ and $T_i$ have a Taylor's series expansion in the real sense.

\begin{prop}
Let $\bT:A\subset\HH\rightarrow\C$ be an operator such that $\bT(\bbf)=\bT(u_{\bbf} + i v_{\bbf}) = \bT(u_{\bbf}, v_{\bbf}) = T_r(u_{\bbf}, v_{\bbf}) + i T_i(u_{\bbf}, v_{\bbf})$.
If the \frechet complex derivative of $\bT$ at a point $\bc\in A$ (i.e., $d\bT(\bc):\HH\rightarrow\C$) exists, then $T_r$ and $T_i$ are differentiable at the point $\bc=(c_1,c_1)=c_1 + i c_2$, where $c_1, c_2 \in \cH$. Furthermore,
\begin{align}\label{EQ:fre_cauchy-riemann}
\nabla_u T_r(c_1,c_2)=\nabla_v T_i(c_1,c_2),\quad
\nabla_v T_r(c_1,c_2)=-\nabla_u T_i(c_1,c_2).
\end{align}
\end{prop}

Equations (\ref{EQ:fre_cauchy-riemann}) are  the \textit{Cauchy Riemann conditions} with respect to the \frechet notion of differentiability. Similar to the simple case of complex valued functions, they provide a necessary and sufficient condition, for a complex operator $\bT$, that is defined on $\HH$, to be differentiable in the complex sense, provided that $\bT$ is differentiable in the real sense. This is explored in the following proposition.

\begin{prop}\label{PRO:fre_CR2}
If  the operator $\bT:A\subseteq\HH\rightarrow\C$, $\bT(\bbf)=T_r(\bbf) + i T_i(\bbf)$, where $\bbf=u_{\bbf} + i v_{\bbf}$, is \frechet differentiable in the real sense at a point $(c_1,c_2)\in\cH^2$ and the \frechet Cauchy-Riemann conditions hold:
\begin{align}
\nabla_u T_r(c_1,c_2)=\nabla_v T_i(c_1,c_2), \quad\nabla_v T_r(c_1,c_2)=-\nabla_u T_i(c_1,c_2),
\end{align}
then $\bT$ is differentiable in the complex sense at the point $\bc=(c_1,c_2)=c_1+c_2 i\in\HH$.
\end{prop}
\begin{proof}
see Appendix \ref{APENDIX:basic_prop}.
\end{proof}

If the \frechet Cauchy Riemann conditions are not satisfied for an operator $\bT$, then the \frechet complex derivative does not exist and the function cannot be expressed in terms of $\bh$, as in the case of \frechet complex differentiable functions (see equation \ref{EQ:fre_complex_Taylor}).  Nevertheless, if $\bT$ is \frechet differentiable in the real sense (i.e., $T_r$ and $T_i$ are \frechet differentiable), we may still find a form of Taylor's series expansion  by utilizing the extension of Wirtinger's calculus. It can be shown (see the proof of proposition \ref{PRO:fre_CR2} in Appendix \ref{APENDIX:basic_prop}), that:
\begin{align}\label{EQ:frechet_wirti2}
\bT(\bc+\bh) =&\bT(\bc) + \frac{1}{2}\left\langle \bh, \left(\nabla_u\bT(\bc) -i\nabla_v\bT(\bc)\right)^*\right\rangle_{\HH}\\
 &+\frac{1}{2}\left\langle \bh^*, \left(\nabla_u\bT(\bc) +i \nabla_v\bT(\bc)\right)^*\right\rangle_{\HH}  + o(\|\bh\|_{\HH}).\nonumber
\end{align}
One may notice that in this case the associated Taylor's expansion is casted in terms of both $\bh$ and $\bh^*$. This can be generalized for higher order Taylor's expansion formulas by following the same rationale. Observe also that, if $\bT$ is \frechet complex differentiable, this relation degenerates (due to the Cauchy Riemann conditions) to the respective Taylor's expansion formula (i.e., (\ref{EQ:fre_complex_Taylor})). In this context, the following definitions come naturally.

We define the \textit{\frechet Wirtinger's gradient} (or \textit{W-gradient} for short) of $\bT$ at $\bc$ as
\begin{align}\label{EQ:fre_wirti_der}
\nabla_{\bbf}\bT(\bc) =& \frac{1}{2}\left(\nabla_1\bT(\bc)  -i \nabla_2\bT(\bc)\right)
= \frac{1}{2}\left(\nabla_u T_r(\bc) + \nabla_v T_i(\bc)\right)\\
      &+ \frac{i}{2}\left(\nabla_u T_i(\bc) - \nabla_v T_r(\bc)\right),\nonumber
\end{align}
and the \textit{\frechet Wirtinger's derivative} (or \textit{$W$-derivative}) as $\frac{\partial \bT}{\partial \bbf}(\bc):\HH\rightarrow\C$, such that $\frac{\partial \bT}{\partial \bbf}(\bc)(\bh)=\langle \bh, \nabla_{\bbf}\bT(\bc)^*\rangle_{\HH}$.
Consequently, the \textit{\frechet conjugate Wirtinger's gradient} (or \textit{CW-gradient} for short) and the \textit{\frechet conjugate Wirtinger's derivative} (or \textit{CW-derivative}) of $\bT$ at $\bc$ are defined by:
\begin{align}\label{EQ:fre_conj_wirti_der}
\nabla_{\bbf^*}\bT(\bc) =& \frac{1}{2}\left(\nabla_1\bT(\bc)  +  i \nabla_2\bT(\bc)\right)
= \frac{1}{2}\left(\nabla_u T_r(\bc) - \nabla_v T_i(\bc)\right)\\
      &+ \frac{i}{2}\left(\nabla_u T_i(\bc) + \nabla_v T_r(\bc)\right),\nonumber
\end{align}
and $\frac{\partial \bT}{\partial \bbf^*}(\bc):\HH\rightarrow\C$, such that $\frac{\partial \bT}{\partial \bbf^*}(\bc)(\bh)=\langle \bh, \left(\nabla_{\bbf^*}\bT(\bc)\right)^*\rangle_{\HH}$.
Note, that both the W-derivative and the CW-derivative
exist, if $\bT$ is \frechet differentiable in the real sense. In view of these new definitions, equation (\ref{EQ:frechet_wirti2}) may now be recasted as follows
\begin{align}\label{EQ:fre_wirti3}
\bT(\bc+\bh) =& \bT(\bc) + \left\langle \bh, \left(\nabla_{\bbf}\bT(\bc)\right)^* \right\rangle_{\HH}
+\left\langle\bh^*, \left(\nabla_{\bbf^*}\bT(\bc)\right)^*\right\rangle_{\HH} + o(\|\bh\|_{\HH}).
\end{align}

From these definitions, several properties can be derived:
\begin{enumerate}
\item If $\bT(\bbf)$ is $\bbf$-holomorphic at $\bc$ (i.e., it has a Taylor series expansion with respect to $\bbf$ at $\bc$), then its \frechet W-derivative at $\bc$ degenerates to the standard \frechet complex derivative and its \frechet CW-derivative vanishes, i.e., $\nabla_{\bbf^*}\bT(\bc)=\bZero$.
\item If $\bT(\bbf)$ is $\bbf^*$-holomorphic at $\bc$ (i.e., it has a Taylor series expansion with respect to $\bbf^*$ at $\bc$), then $\nabla_{\bbf}\bT(\bc)=\bZero$.
\item The first order Taylor expansion around $\bbf\in\HH$ is given by
\begin{align*}
\bT(\bbf+\bh) =& \bT(\bbf) + \langle \bh, \left(\nabla_{\bbf} \bT(\bbf)\right)^* \rangle_\HH
+ \langle \bh^*, \left(\nabla_{\bbf^*} \bT(\bbf)\right)^* \rangle_\HH.
\end{align*}
\item If $\bT(\bbf)=\langle \bbf, \bw\rangle_\HH$, then $\nabla_{\bbf}\bT(\bc)=\bw^*$, $\nabla_{\bbf^*}\bT(\bc)=\bZero$, for every $\bc$.
\item If $\bT(\bbf)=\langle \bbf^*, \bw\rangle_\HH$, then $\nabla_{\bbf}\bT(\bc)=\bZero$, $\nabla_{\bbf^*}\bT(\bc)=\bw^*$, for every $\bc$.
\item Linearity: If $\bT,\bS:\HH\rightarrow\C$ are \frechet differentiable in the real sense at $\bc\in\HH$ and $\alpha, \beta\in\C$, then
\begin{align*}
\nabla_{\bbf}(\alpha \bT + \beta \bS)(\bc) &= \alpha\nabla_{\bbf}\bT(\bc) + \beta\nabla_{\bbf}\bS(\bc)\\
\nabla_{\bbf^*}(\alpha \bT + \beta \bS)(\bc) &= \alpha\nabla_{\bbf^*}\bT(\bc) + \beta\nabla_{\bbf^*}\bS(\bc).
\end{align*}
\end{enumerate}
A complete list of the derived properties, together with the proofs of the most important ones, are given in Appendix \ref{APPENDIX:wirti_properties}.

An important consequence of the previous properties is that if $T$ is a real valued operator defined on $\HH$, then $\left(\nabla_{\bbf} T(\bc)\right)^* = \nabla_{\bbf^*} T(\bc)$, and its first order Taylor's expansion is given by:
\begin{align*}
T(\bbf+\bh) & =  T(\bbf) + \langle \bh, \left(\nabla_{\bbf}\bT(\bbf)\right)^*\rangle_\HH + \langle \bh^*, \left(\nabla_{\bbf^*}T(\bbf)\right)^* \rangle_\HH\\
& = T(\bbf) + \langle \bh, \nabla_{\bbf^*}\bT(\bbf)\rangle_\HH + \left(\langle \bh, \nabla_{\bbf^*}T(\bbf)\rangle_\HH\right)^*
 = T(\bbf) + 2\cdot \Re\left[ \langle \bh, \nabla_{\bbf^*}T(\bbf)\rangle_\HH\right].
\end{align*}
However, in view of the Cauchy Riemann inequality we have:
\begin{align*}
\Re\left[ \langle \bh, \nabla_{\bbf^*}T(\bbf)\rangle_\HH\right] & \leq \left|\langle \bh, \nabla_{\bbf^*}T(\bbf)\rangle_\HH\right|
\leq \|\bh\|_\HH \cdot \| \nabla_{\bbf^*}T(\bbf)\|_\HH.
\end{align*}
The equality in the above relationship holds if $\bh\upuparrows \nabla_{\bbf^*}T$ (where the notation $\upuparrows$ denotes that $\bh$ and $\nabla_{\bbf^*}T$  have the same direction, i.e., there is a $\lambda>0$, such that $\bh=\lambda \nabla_{\bbf^*}T$). Hence, the direction of increase of $T$ is $\nabla_{\bbf^*}T(\bbf)$. Therefore, any gradient descent based algorithm minimizing $T(\bbf)$ is based on the update scheme:
\begin{align}
\bbf_{n} = \bbf_{n-1} - \mu\cdot\nabla_{\bbf^*}T(\bbf_{n-1}).
\end{align}

Assuming differentiability of $T$, a standard result from \frechet real calculus states that a necessary condition for a point $\bc$ to be an optimum (in the sense that $T(\bbf)$ is minimized or maximized) is that this point is a stationary point of $T$, i.e., the \frechet partial derivatives of $T$ at $\bc$ vanish. In the context of Wirtinger's calculus, we have the following obvious corresponding result.

\begin{prop}\label{PRO:fre_first_order_opt}
If the function  $T:A\subseteq\HH\rightarrow\R$ is \frechet differentiable at $\bc$ in the real sense, then a necessary condition for a point $\bc$ to be a local optimum (in the sense that $T(\bc)$ is minimized or maximized) is that either the \frechet W, or the \frechet CW derivative vanishes.
\end{prop}

\begin{proof}
Observe that if $T$ is real valued, the Wirtinger derivatives take the form $\nabla_{\bbf}T(\bc)=\frac{1}{2}(\nabla_u T(\bc) - i\nabla_v T(\bc))$ and $\nabla_{\bbf^*}T(\bc)=\frac{1}{2}(\nabla_u T(\bc) + i\nabla_v T(\bc))$. If $\bc$ is a local optimum of $T$ then $\nabla_u T(\bc) = \nabla_v T(\bc) = 0$ and thus $\nabla_{\bbf}T(\bc)=\nabla_{\bbf^*}T(\bc)=0$. Note, that for real valued functions the W and the CW derivatives constitute a conjugate pair. Thus, if the W derivative vanishes, then the CW derivative vanishes too. The converse is also true. This completes the proof.
\end{proof}

%--------------------------------------------------------------------------------
\section{Complex Kernel Least Mean Squares - CKLMS}\label{SEC:CKLMS}
%--------------------------------------------------------------------------------
In order to illustrate how the proposed framework may be applied to problems of complex signal processing, we present two realizations of the Kernel Least Mean Squares (KLMS) algorithm for complex data. The first scheme (CKLMS1) employs the complexification of real reproducing kernels (see section \ref{SEC:Complexification}), while the second one uses pure complex kernels (CKLMS2). Wirtinger's calculus is exploited in both cases to compute the necessary gradient updates.

%--------------------------------------------------------------------------------
\subsection{Complex KLMS via complexification of real kernels - CKLMS1}\label{SEC:CKLMS1}
Consider the sequence of examples $(\bz(1),d(1))$, $(\bz(2),d(2))$, $\dots$, $(\bz(N),d(N))$, where $d(n)\in\C$, $\bz(n)\in V\subset\C^\nu$, $\bz(n)=\bx(n) + i \by(n)$, $\bx(n), \by(n)\in\R^\nu$, for $n=1,\dots,N$. Consider, also, a real reproducing kernel $\kappa$ defined on $X\times X$, $X\subseteq\R^{2\nu}$, and let $\cH$ be the corresponding RKHS. We map the points $\bz(n)$ to the RKHS $\HH$ ($\HH$ is constructed as explained in section \ref{SEC:Complexification}) using the mapping $\bPhi$:
\begin{align*}
\bPhi(\bz(n)) &= \Phi(\bz(n)) + i \Phi(\bz(n)) = \kappa\left(\cdot, (\bx(n),\by(n))\right) + i\cdot\kappa\left(\cdot, (\bx(n),\by(n))\right),
\end{align*}
for $n=1,\dots,N$, where $\Phi$ is the feature map of $\cH$. The objective of the complex Kernel LMS is to design a filter, $\bw$, with desired response $\hat d(n)=\langle \bPhi(\bz(n)), \bw\rangle_\HH$, so that to minimize $E\left[\cL_n(\bw)\right]$, where
\begin{align*}
\cL_n&(\bw) =  |e(n)|^2 = |d(n) - \langle \bPhi(\bz(n)), \bw\rangle_\HH|^2\\
&= \left(d(n) - \langle \bPhi(\bz(n)), \bw\rangle_\HH\right) \left(d(n) - \langle \bPhi(\bz(n)), \bw\rangle_\HH\right)^*\\
&= \left(d(n) - \langle \bw^*, \bPhi^*(\bz(n))\rangle_\HH\right) \left(d(n)^* - \langle \bw, \bPhi(\bz(n))\rangle_\HH\right),
\end{align*}
at each instance $n$. We then apply the complex LMS to the transformed data, estimating the mean square error by its current measurement $\hat E\left[\cL_n(\bw)\right] = \cL_n(\bw)$, using the rules of Wirtinger's calculus to compute the CW gradient, i.e., $\nabla_{\bw^*}\cL_n(\bw)=-e(n)^*\cdot\bPhi(\bz(n))$.
Therefore the CKLMS1 update rule becomes:
\begin{align}
\bw(n) = \bw(n-1) + \mu e(n)^*\cdot\bPhi(\bz(n)),
\end{align}
where $\bw(n)$ denotes the estimate at iteration $n$.

Assuming that $\bw(0)=\bZero$, the repeated application of the weight-update equation gives:
\begin{align}
\bw(n) = & \bw(n-1) + \mu e(n)^*\bPhi(\bz(n))
=  \bw(n-2) + \mu e(n-1)^*\bPhi(\bz(n-1))
 + \mu e(n)^*\bPhi(\bz(n))\nonumber\\
= & \mu \sum_{k=1}^{n} e(k)^*\bPhi(\bz(k))\label{EQ:CKLMS_W}.
\end{align}
Thus, the filter output at iteration $n$ becomes:
\begin{align}
\hat d(n) =&\langle \bPhi(\bz(n)), \bw(n-1) \rangle_\HH
= \mu \sum_{k=1}^{n-1} e(k) \langle \bPhi(\bz(n)), \bPhi(\bz(k)) \rangle_\HH\nonumber\\
=& 2\mu \sum_{k=1}^{n-1} \Re[e(k)]\kappa(\bz(n), \bz(k))
+ 2\mu \cdot i \sum_{k=1}^{n-1} \Im[e(k)]\kappa(\bz(n), \bz(k)), \label{EQ:NCKLMS1_des_resp}
\end{align}
where the evaluation of the kernel is done by replacing the complex vectors $\bz(n)$, of $\C^\nu$ with the corresponding real vectors of $\R^{2\nu}$, i.e., $\bz(n)= (\bx(n),\by(n))$.

It can readily be shown that, since the CKLMS1 is the complex LMS in RKHS, the important properties of the LMS (convergence in the mean, misadjustment, e.t.c.) carry over to CKLMS1. Furthermore, we may also define a normalized version, which we call \textit{Normalized Complex Kernel LMS} (NCKLMS1). The weight-update of the NCKLMS1 is given by:
\begin{align*}
\bw(n) = & \bw(n-1) + \frac{\mu}{2\cdot\kappa(\bz(n),\bz(n))} e(n)^*\bPhi(\bz(n))
\end{align*}
The NCKLMS1 algorithm is summarized in Algorithm \ref{ALG:NCKLMS1}. We should emphasize that this formulation of the complex KLMS cannot be derived following the usual ``black box'' rationale of the kernel trick, as it has already been pointed out in section \ref{SEC:Complexification}. The complexified real kernel trick can be used instead.

\begin{algorithm}
\caption{Normalized Complex Kernel  LMS with complexification of real kernels (NCKLMS1)}\label{ALG:NCKLMS1}
\textbf{INPUT: } $(\bz(1),d(1))$, $\dots$, $(\bz(N),d(N))$\\
\textbf{OUTPUT:} The expansion \\
$\bw=\sum_{k=1}^{N} a(k)\kappa(\cdot, \bz(k)) + i\cdot \sum_{k=1}^{N} b(k)\kappa(\cdot,\bz(k))$.\\
\\
\textbf{Initialization:} Set  $\ba=\{\}$, $\bb=\{\}$, $\bZ=\{\}$ (i.e., $\bw=\bZero$). Select the step parameter $\mu$ and the kernel $\kappa$.
\begin{algorithmic}
\FOR{$n=1:N$}
\STATE{Compute the filter output:
\begin{align*}
\hat d(n) =& \sum_{k=1}^{n-1}(a(k)+b(k))\cdot\kappa(\bz(n),\bz(k))
     + i \sum_{k=1}^{n-1}(a(k)-b(k))\cdot\kappa(\bz(n),\bz(k)).
\end{align*}}
\STATE{Compute the error: $e(n)=d(n)-\hat d(n)$.}
\STATE{$\gamma=2\kappa(\bz(n),\bz(n))$.}
\STATE{$a(n)=\mu(\Re[e(n)] + \Im[e(n)])/\gamma$.}
\STATE{$b(n)=\mu(\Re[e(n)] - \Im[e(n)])/\gamma$.}
\STATE{Add the new center $\bz(n)$ to the list of centers, i.e., add $\bz(n)$ to the list $\bZ$, add $a(n)$ to the list $\ba$, add $b(n)$ to the list $\bb$.}
\ENDFOR
\end{algorithmic}
\end{algorithm}

One might think, that modeling the desired response as $\hat d(n)=\langle \bw(n-1), \bPhi(\bz(n))\rangle_{\HH}$, provides an alternative formulation for the CKLMS1 algorithm. In this case, the CW gradient of the instantaneous square error is given by $\nabla_{\bw^*}\cL_n(\bw) = -e(n)\bPhi(\bz(n))$. Following the same procedure, we conclude that the update rule becomes:
$\bw(n) = \bw(n-1) + \mu e(n)\cdot\bPhi(\bz(n))$,
and assuming that $\bw(0)=\bZero$, one concludes that:
\begin{align*}
\bw(n) = & \mu \sum_{k=1}^{n} e(k)\bPhi(\bz(k)).
\end{align*}
However, although this relation is different to equation (\ref{EQ:CKLMS_W}), the filter output at iteration $n$, for this filter, turns out to be exactly the same as before:
\begin{align*}
\hat d(n) =&\langle \bw(n-1), \bPhi(\bz(n)) \rangle_\HH
= \mu \sum_{k=1}^{n-1} e(k) \langle \bPhi(\bz(k)), \bPhi(\bz(n)) \rangle_\HH,
\end{align*}
which is in line with what we know for the standard complex LMS.

%--------------------------------------------------------------------------------
\subsection{Complex KLMS with pure complex kernels - CKLMS2}\label{SEC:CKLMS2}
As, in section \ref{SEC:CKLMS1}, consider the sequence of examples $(\bz(1),d(1))$, $(\bz(2),d(2))$, $\dots$, $(\bz(N),d(N))$, where $d(n)\in\C$, $\bz(n)\in V\subset\C^\nu$, $\bz(n)=\bx(n) + i \by(n)$, $\bx(n), \by(n)\in\R^\nu$, for $n=1,\dots,N$. Consider also a complex reproducing kernel $\kappa$ defined on $X\times X$, $X\subseteq\C^\nu$ and the respective complex RKHS $\HH$. Each element $\bbf\in\HH$ may be cast in the form $\bbf=u_{\bbf} + i v_{\bbf}$, $u_{\bbf}, v_{\bbf}\in\cH$, where $\cH$ is a real Hilbert space. We map the points $\bz(n)$ to the complex RKHS $\HH$ using the feature map  $\tilde\bPhi:\XX\rightarrow\HH: \tilde\bPhi(\bz)=\langle\cdot, \kappa(\cdot,\bz)\rangle_{\HH}$,
for $n=1,\dots,N$. Estimating the filter output by $\hat d(n)=\langle\tilde\bPhi(\bz(n)), \bw\rangle_{\HH}$, the objective of the complex Kernel LMS is to minimize $E\left[\cL_n(\bw)\right]$, at each instance $n$. Once more, we apply the complex LMS to the transformed data, using the rules of Wirtinger's calculus to compute the gradient of $\cL_n(\bw)$, i.e., $\nabla_{\bw^*}\cL_n(\bw)=-e(n)^*\cdot\tilde\bPhi(\bz(n))$.
Therefore, the CKLMS2 update rule becomes $\bw(n) = \bw(n-1) + \mu e(n)^*\cdot\tilde\bPhi(\bz(n))$,  as expected, where $\bw(n)$ denotes the estimate at iteration $n$.

Assuming that $\bw(0)=\bZero$, the repeated application of the weight-update equation gives:
\begin{align}
\bw(n) =  \sum_{k=1}^{n} e(k)^*\tilde\bPhi(\bz(k))\label{EQ:CKLMS2_W}.
\end{align}
Thus, the filter output at iteration $n$ becomes:
\begin{align*}
\hat d(n) =&\langle \tilde\bPhi(\bz(n)), \bw(n-1) \rangle_\HH
= \mu \sum_{k=1}^{n-1} e(k) \langle \tilde\bPhi(\bz(n)), \tilde\bPhi(\bz(k)) \rangle_\HH
= \mu \sum_{k=1}^{n-1} e(k) \kappa(\bz(k), \bz(n)).
\end{align*}

We should note, that the CKLMS2 algorithm may be equivalently derived, if one blindly applies the kernel trick on the complex LMS. However, such an approach conceals the mathematical framework that lies underneath, which is needed if one seeks a deeper understanding of the problem. The repeated application of the update equation of the CLMS yields:
\begin{align*}
\bw(n) =  \sum_{k=1}^{n} e(k)^*\bz(k),
\end{align*}
while the filter output at iteration $n$ is given by:
\begin{align*}
\hat d(n) = \mu \sum_{k=1}^{n-1} e(k) \bz(n)^H\bz(k),
\end{align*}
where the notation $\cdot^H$ denotes the Hermitian matrix. It is evident that the application of the kernel trick on these equations yields the same results.

Furthermore, note that, using the complex gaussian kernel, the algorithm is automatically normalized.
The CKLMS2 algorithm is summarized in Algorithm \ref{ALG:NCKLMS2}.

Another formulation of the CKLMS2 algorithm may be derived if we estimate the filter output as $\hat d(n) = \langle \tilde\bw(n-1), \bPhi(\bz(n)) \rangle_{\HH}$. Then the update rule becomes
\begin{align*}
\bw(n) = \bw(n-1) + \mu e(n)\cdot\tilde\Phi(\bz(n)).
\end{align*}
Assuming that $\bw(0)=\bZero$, the repeated application of the weight-update equation gives:
\begin{align*}
\bw(n) =  \sum_{k=1}^{n} e(k)\tilde\bPhi(\bz(k)),
\end{align*}
and the filter output at iteration $n$ becomes:
\begin{align}
\hat d(n) =  \mu \sum_{k=1}^{n-1} e(k) \kappa(\bz(n), \bz(k)). \label{EQ:NCKLMS2_des_resp}
\end{align}
Note that the two formulations of the CKLMS2 are not identical, as it was the case for CKLMS. However, all the simulated experiments that we performed, using the complex gaussian kernel, exhibited similar performance (in terms of signal to noise ratio - SNR).

\begin{algorithm}
\caption{Normalized Complex Kernel  LMS2 (NCKLMS2)}\label{ALG:NCKLMS2}
\textbf{INPUT: } $(\bz(1),d(1))$, $\dots$, $(\bz(N),d(N))$\\
\textbf{OUTPUT:} The expansion \\
$\bw=\sum_{k=1}^{N} a(k)\kappa(\cdot,\bz(k))$.\\
\\
\textbf{Initialization:} Set  $\ba=\{\}$, $\bZ=\{\}$ (i.e., $\bw=\bZero$). Select the step parameter $\mu$ and the parameter $\sigma$ of the complex gaussian kernel.
\begin{algorithmic}
\FOR{$n=1:N$}
\STATE{Compute the filter output:
$$\hat d(n) = \sum_{k=1}^{n-1}a(k)\cdot\kappa(\bz(k),\bz(n)).$$}
\STATE{Compute the error: $e(n)=d(n)-\hat d(n)$.}
\STATE{$\gamma=\kappa(\bz(n),\bz(n))$.}
\STATE{$a(n)=\mu e(n) / \gamma$.}
\STATE{Add the new center $\bz(n)$ to the list of centers, i.e., add $\bz(n)$ to the list $\bZ$, add $a(n)$ to the list $\ba$.}
\ENDFOR
\end{algorithmic}
\end{algorithm}

%--------------------------------------------------------------------------------
\subsection{Sparsification}\label{SEC:sparsif}

The main drawback of any kernel based adaptive filtering algorithm is that a growing number of training points, $\bz(n)$, is involved, as it is apparent from (\ref{EQ:CKLMS_W}), (\ref{EQ:CKLMS2_W}) in the case of complex KLMS. Hence, increasing memory and computational resources are needed, as time evolves. Several strategies have been proposed to cope with this problem and to come up with sparse solutions. In this paper, we employ the well known \textit{novelty criterion} \cite{Platt, LiuPriHay}.
In novelty criterion online sparsification, a dictionary of points, C, is formed and updated appropriately. Whenever a new data pair $(\bPhi(\bz_n),d_n)$ is considered, a decision is immediately made of whether to add the new center, $\bPhi(\bz(n))$, to the dictionary of centers $\cC$. The decision is reached following two simple rules. First, the distance of the new center, $\bPhi(\bz(n))$, from the current dictionary is evaluated: $dis = \min_{\bc_k\in\cC}\{\|\bPhi(\bz(n)) - \bc_k\|_\HH\}$.
If this distance is smaller than a given threshold $\delta_1$ (i.e., the new center is close to the existing dictionary), then the center is not added to $\cC$. Otherwise, we compute the prediction error $e_n = d_n - \hat d_n$. If $|e_n|$ is smaller than a predefined threshold $\delta_2$, then the new center is discarded. Only if $|e_n|\geq\delta_2$ the new center $\bPhi(\bz(n))$ is added to the dictionary.

An alternative method has been considered in \cite{SlaTheYam}, which results in an exponential forgetting mechanism of past data. In \cite{VanVaerenbergh.Sliding.RLS, SlaThe}, the sliding window rationale has been considered. In all the implementations of CKLMS that are presented in this paper the novelty criterion was adopted.

%-----------------------------------
%\subsection{Relation with complex RBF networks}
%Observing the filter output expansions (\ref{EQ:NCKLMS1_des_resp}) and (\ref{EQ:NCKLMS2_des_resp}), one may notice the %similarity of the proposed methods with complex RBF networks (see \cite{Chen91}, \cite{Chen94}, \cite{Chen08}). Although not %identical, the filter output expansions of both methods bear a great resemblance. However, the proposed kernel-based methods %are significantly more robust, as they ensure convergence to a global minimum and do not suffer from the well known problems %that occur in neural networks (trapping to a local minimum e.t.c.). Moreover, the celebrated Representer Theorem %(\cite{KimWahba}, \cite{SchoSmo}, \cite{TheoKou}) ensures that the solution is optimal. These improved characteristics of %kernel-based methods over RBF networks are well documented in the related literature \cite{TheoKou}.

\begin{figure}
\begin{center}
\includegraphics[scale=0.35]{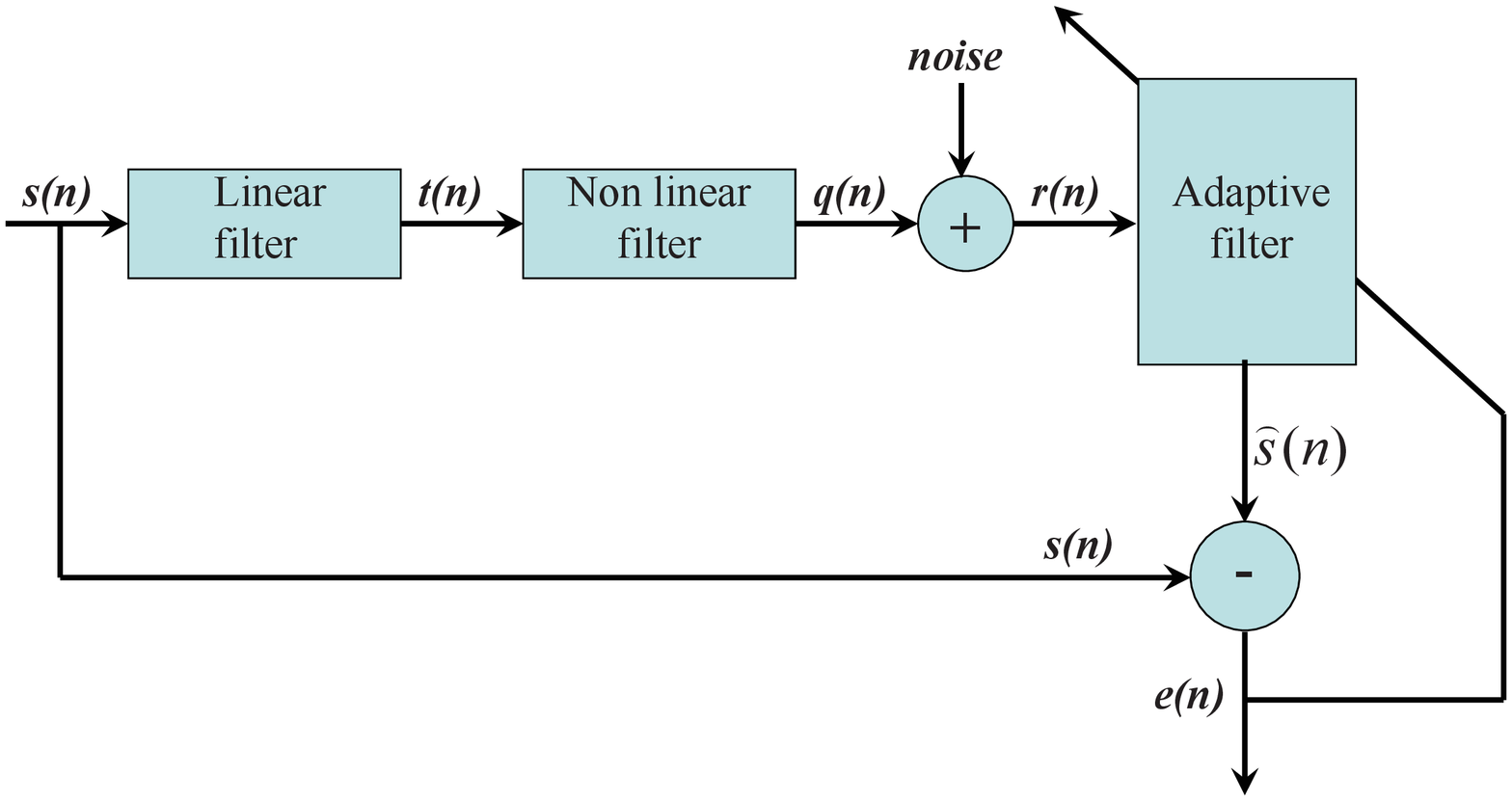}
\end{center}
\caption{The equalization task.}\label{FIG:equal_form}
\end{figure}

%--------------------------------------------------------------------------------
\section{Experiments}\label{SEC:Experim}
%--------------------------------------------------------------------------------
The performances of CKLMS1 and CKLMS2 have been tested in the context of: a) a nonlinear channel equalization task (see figure \ref{FIG:equal_form}) and b) a nonlinear channel identification task.

%----------------------------------------
\subsection{Channel Equalization}
For the first case, two nonlinear channels have been considered. The first channel (labeled as \textit{soft nonlinear channel} in the figures) consists of a linear filter:
\begin{align*}
t(n)= (-0.9+0.8i)\cdot s(n) + (0.6-0.7i)\cdot s(n-1)
\end{align*}
and a memoryless nonlinearity
\begin {align*}
q(n) =&\; t(n) + (0.1+0.15i)\cdot t^2(n)
  + (0.06+0.05i)\cdot t^3(n).
\end{align*}
The second one (labeled as \textit{strong nonlinear channel} in the figures) is comprised by the same linear filter and the nonlinearity:
\begin {align*}
q(n) =&\; t(n) + (0.2+0.25i)\cdot t^2(n)
  + (0.12+0.09i)\cdot t^3(n).
\end{align*}
These are standard models that have been extensively used in the literature for such tasks \cite{LiPokPrin}.
At the receiver end of the channels, the signal is corrupted by white Gaussian noise and then observed as $r(n)$. The level of the noise was set to 16dB. The input signal that was fed to the channels had the form
\begin{align}\label{EQ:input}
s(n) = 0.70\left(\sqrt{1-\rho^2}X(n) + i\rho Y(n)\right),
\end{align}
where $X(n)$ and $Y(n)$ are gaussian random variables. This input is circular for $\rho=\sqrt{2}/2$ and highly non-circular if $\rho$ approaches 0 or 1 \cite{Adali10}. Note that the issue of circularity is very important in complex adaptive filtering. Circularity is intimately related to rotation in the geometric sense. A complex random variable $Z$ is called circular, if for any angle $\phi$ both $Z$ and $Z e^{i\phi}$ (i.e., the rotation of $Z$ by angle $\phi$) follow the same probability distribution \cite{ManGoh}. Loosely speaking, non circularity adds some form of nonlinearity to the signal. It can be proved that widely linear estimation (i.e., linear estimation in both $z$ and $z^*$) outperforms standard linear estimation for general (i.e., circular or non-circular) complex signals. For circular signals, the two models lead to identical results \cite{Picin94, Picin95}.

The aim of a channel equalization task is to construct an inverse filter, which acts on the output $r(n)$ and reproduces the original input signal as close as possible. To this end, we apply the NCKLMS1 and the NCKLMS2 algorithms to the set of samples
\begin{align*}
\left((r(n+D), r(n+D-1), \dots, r(n+D-L+1)), s(n)\right),
\end{align*}
where $L>0$ is the filter length and $D$ the equalization time delay, which is present to, almost, any equalization set up.

\begin{figure}
\begin{center}
\includegraphics[scale=0.5]{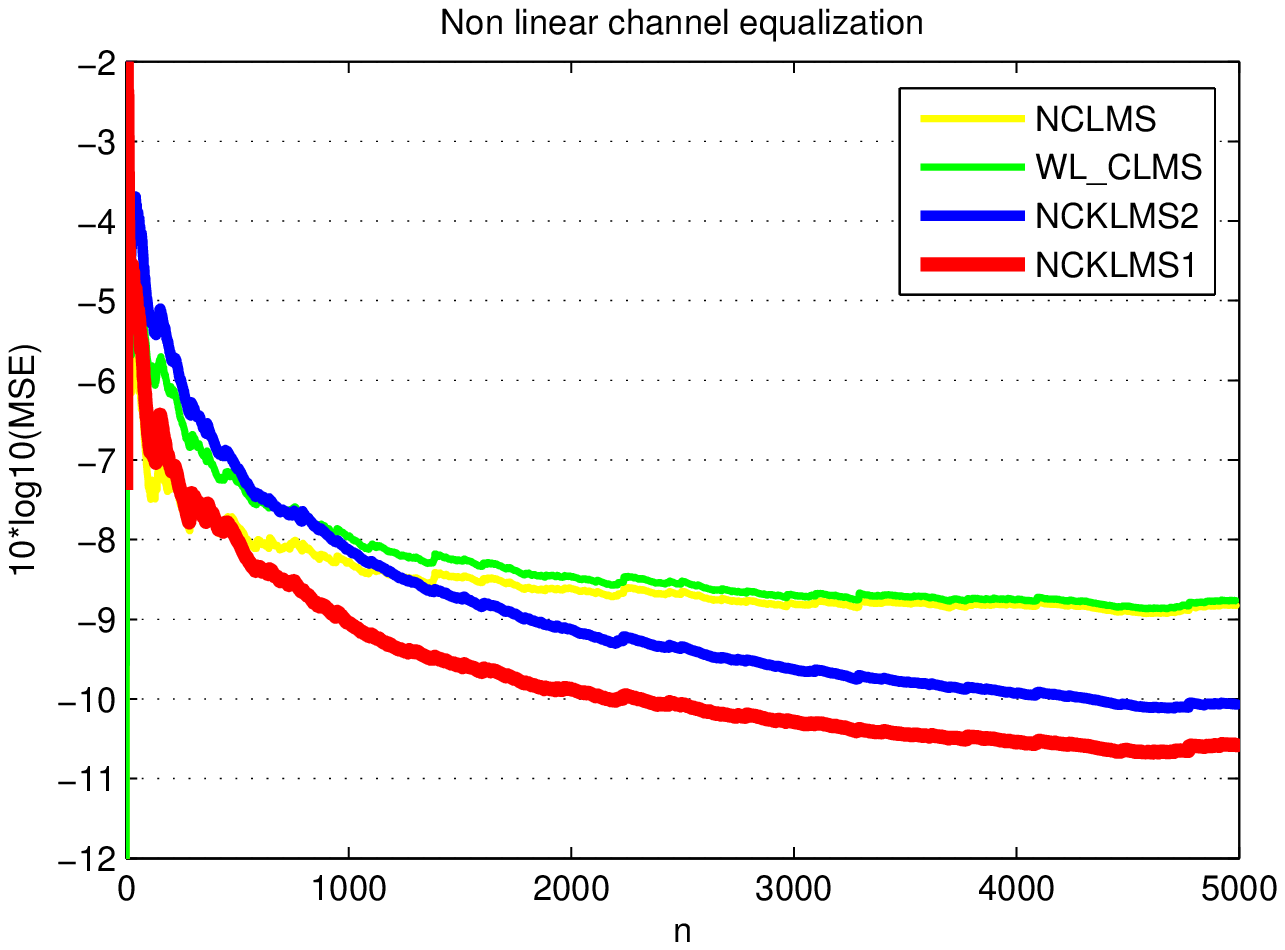}
\includegraphics[scale=0.5]{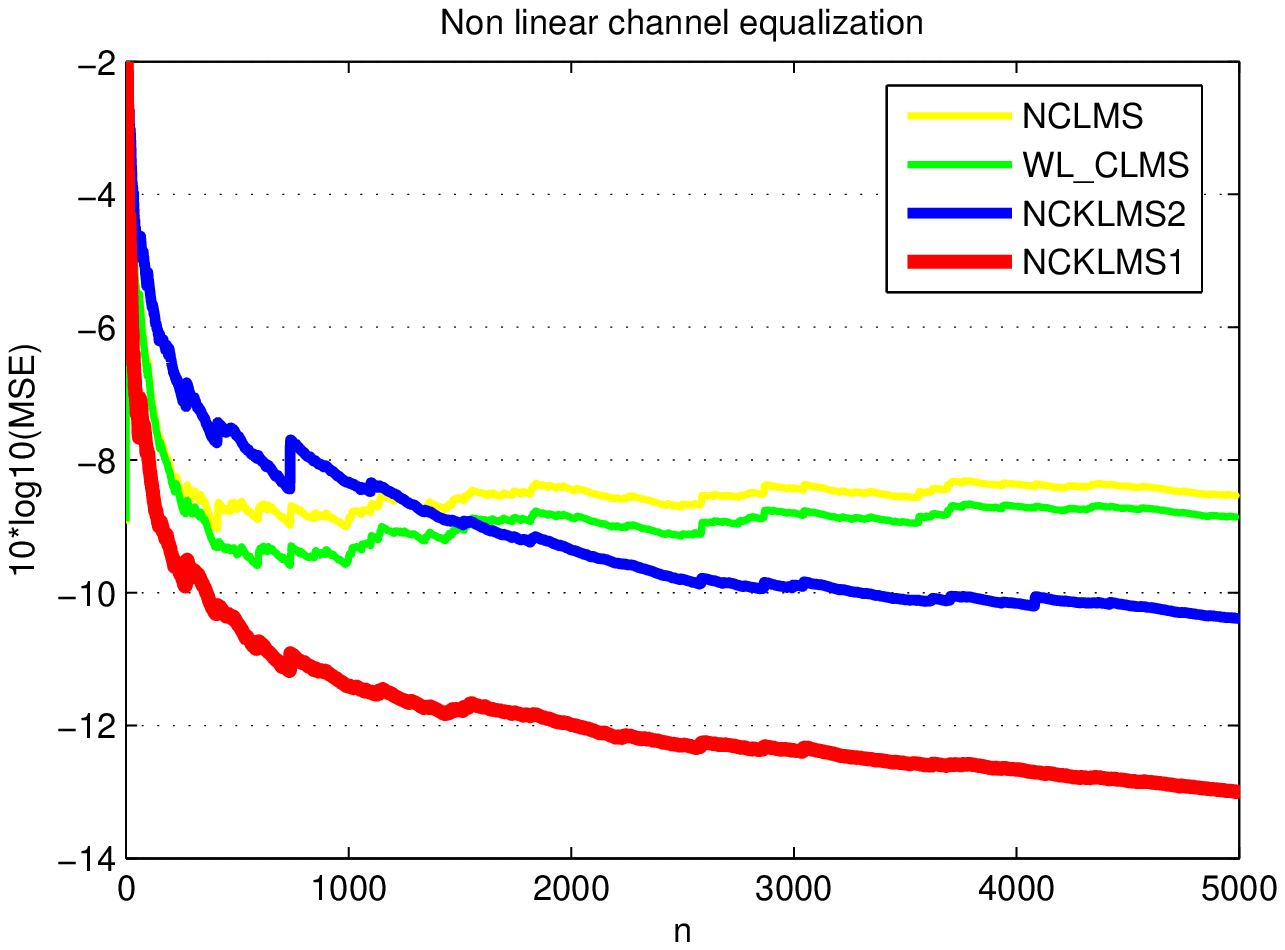}\\
(a)\hspace{18em}(b)
\end{center}
\caption{Learning curves for KÍCLMS1 ($\mu=1/2$), KÍCLMS2, ($\mu=1/4$), ÍCLMS ($\mu=1/16$) and WL-ÍCLMS ($\mu=1/16$) (filter length $L=5$, delay $D=2$) for the soft nonlinear channel equalization problem, for (a) the circular input case, (b) the non-circular input case ($\rho=0.1$).}\label{FIG:equal_circu}
\end{figure}

\begin{figure}
\begin{center}
\includegraphics[scale=0.5]{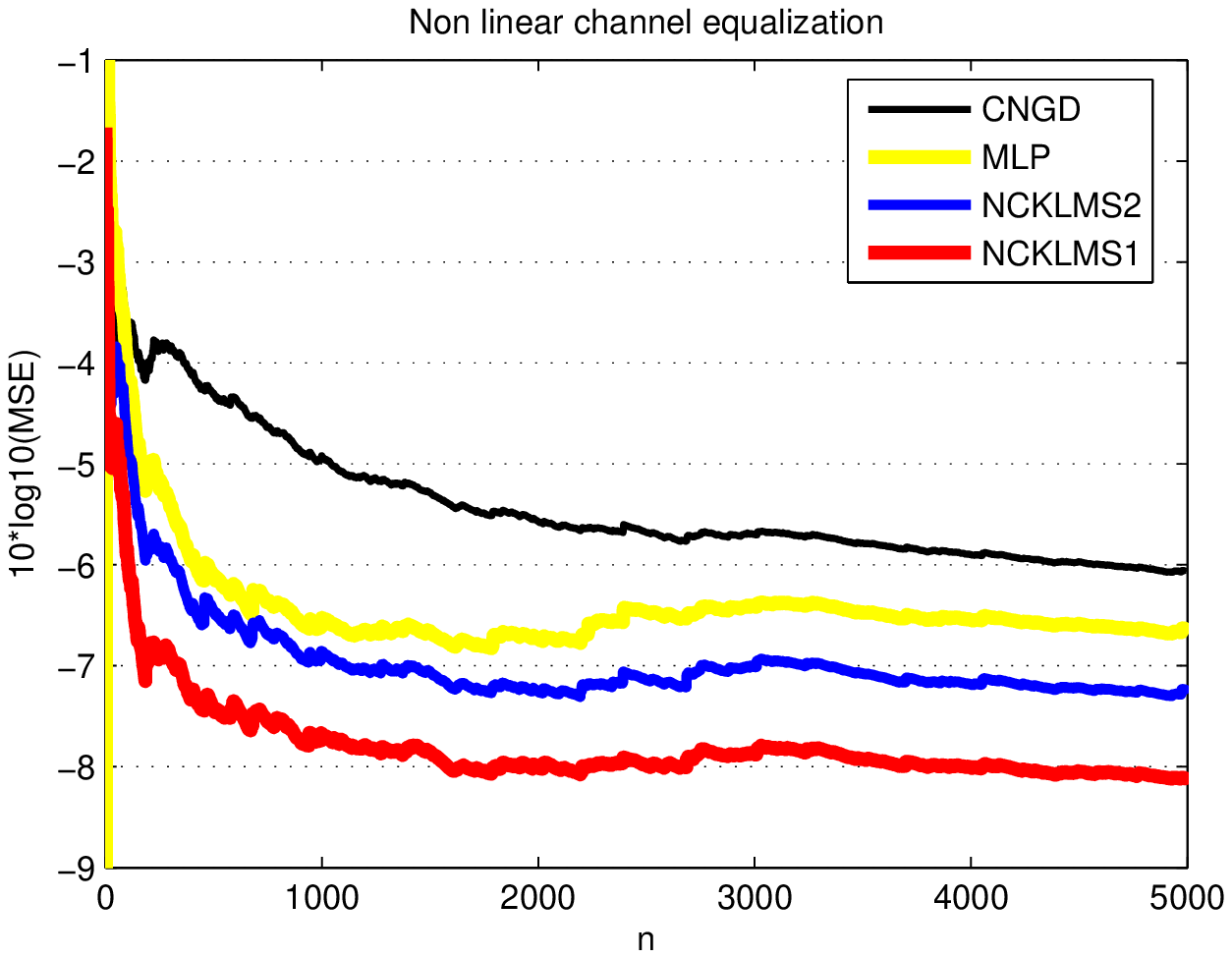}
\includegraphics[scale=0.5]{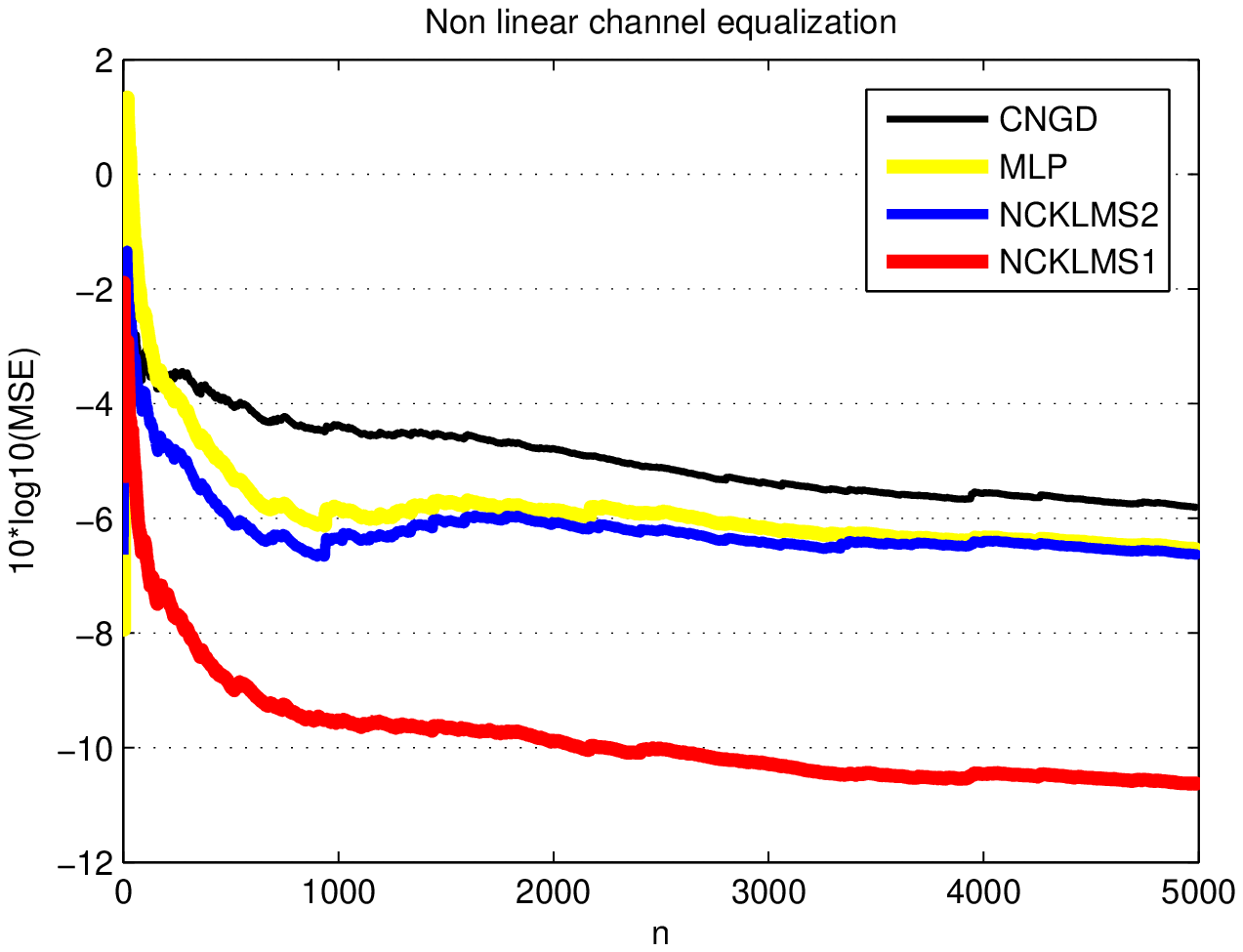}\\
(a) \hspace{18em}(b)
\end{center}
\caption{Learning curves for KÍCLMS1 ($\mu=1/2$), KÍCLMS2, ($\mu=1/4$), CNGD and $L$-50-1 MLP (filter length $L=5$, delay $D=2$) for the hard nonlinear channel equalization problem,  for (a) the circular input case, (b) the non-circular input case ($\rho=0.1$).}\label{FIG:equal_circu2}
\end{figure}

\begin{figure}
\begin{center}
\includegraphics[scale=0.5]{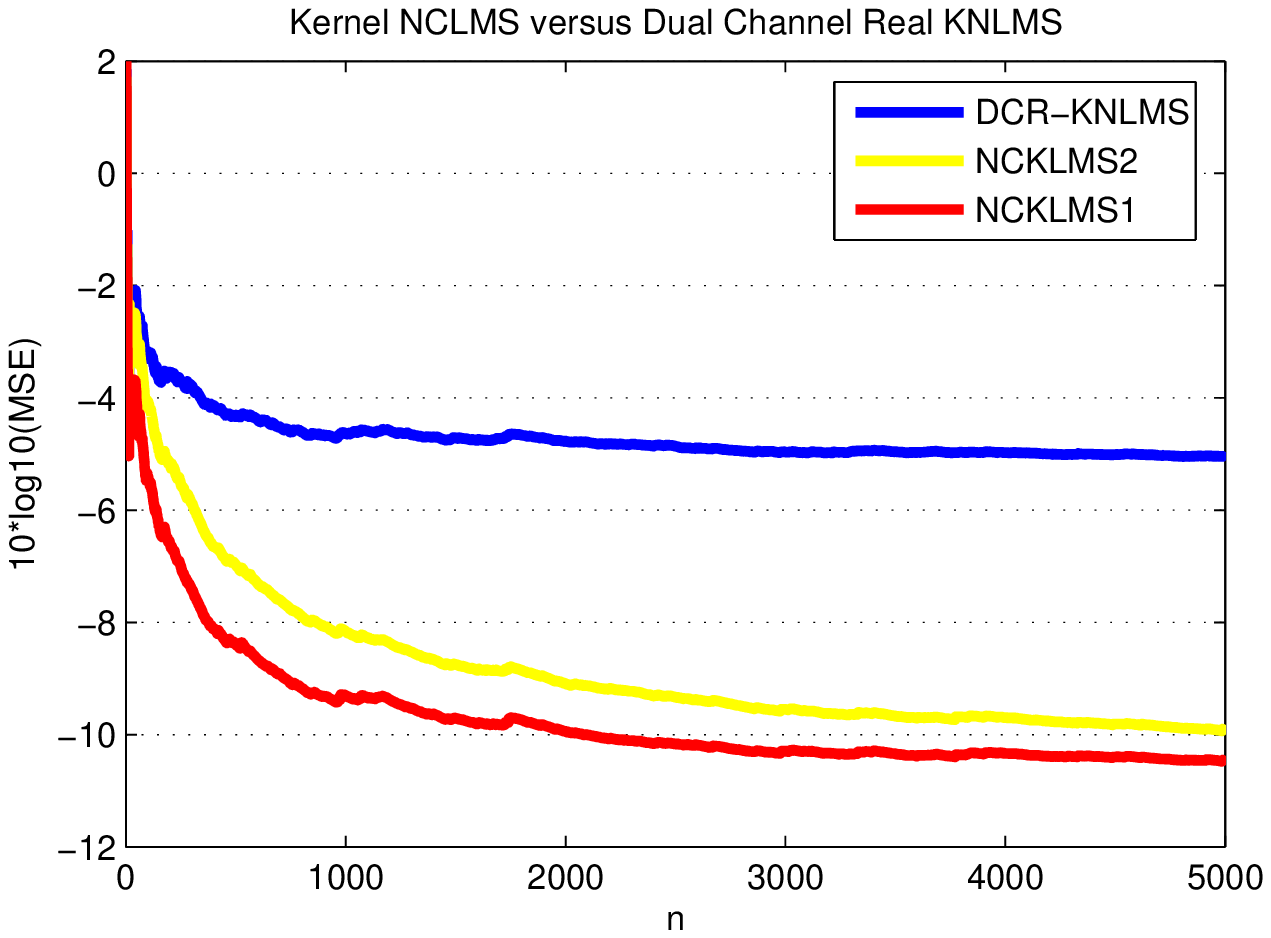}
\includegraphics[scale=0.5]{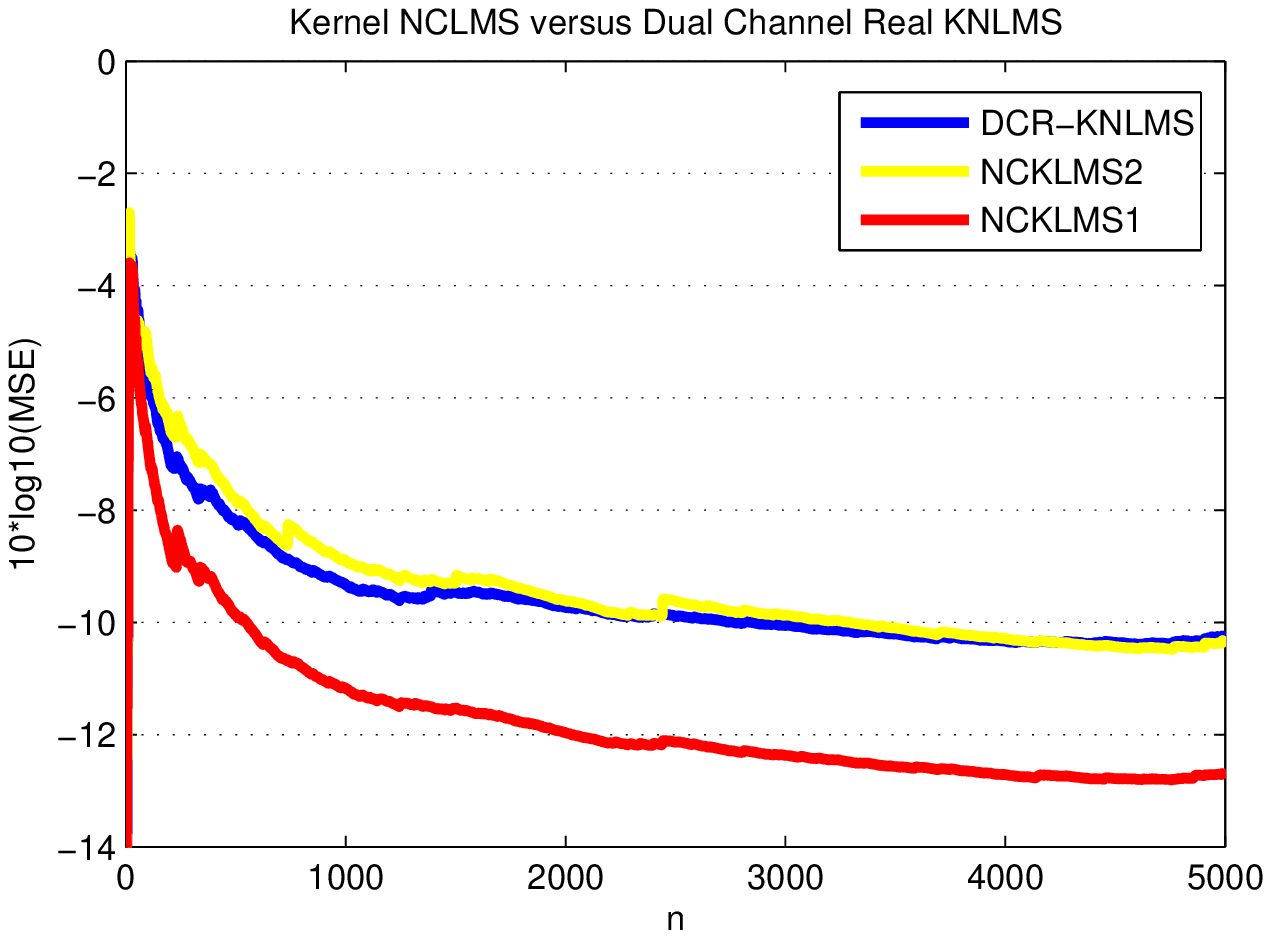}\\
(a) \hspace{18em}(b)
\end{center}
\caption{Learning curves for KÍCLMS1 ($\mu=1/2$) and Dual Channel Real KLMS ($\mu=1/2$)  for the soft nonlinear channel equalization problem, for (a) the circular input case, (b) the non-circular input case ($\rho=0.1$).}\label{FIG:equal_split}
\end{figure}

Experiments were conducted on a set of 5000 samples of the input signal (\ref{EQ:input}) considering both the circular and the non-circular cases. The results are compared with the NCLMS and the WL-NCLMS (i.e., widely linear NCLMS) algorithms and with two adaptive nonlinear algorithms: a) the CNGD algorithm, which is thoroughly described in \cite{ManGoh} and a Multi Layer Perceptron (MLP) with 50 nodes in the hidden layer (proposed in \cite{Adali10}). In both cases, the complex $\tanh$ activation function was employed. Note that the WL-NCLMS has been recently used as an alternative to the CLMS, in an attempt to cope with non circularity as well as with soft nonlinearities. In all algorithms, the step update parameter, $\mu$, is tuned for the best possible results (in terms of the steady-state error rate). For the case of the MLP, the design was also tuned so that the best possible results were obtained.  Time delay $D$ was also set for optimality. Figure \ref{FIG:equal_circu}, shows the learning curves of the NCKLMS1 using the real Gaussian kernel $\kappa(\bx,\by)=\exp(-\|\bx-\by\|^2/\sigma^2)$ (with $\sigma=5$) and the NCKLMS2 using the complex Gaussian kernel $\kappa_{\sigma,\C^d}(\bz,\bw) : = \exp\left(-\frac{\sum_{i=1}^{d}(z_i-w_i^*)^2}{\sigma^2}\right)$ (with $\sigma=5$), together with those obtained from the NCLMS and the WL-NCLMS algorithms. Figure \ref{FIG:equal_circu2}  shows the learning curves of the NCKLMS1 and NCKLMS2 versus the CNGD and the $L$-50-1 MLP. Finally, figure \ref{FIG:equal_split} compares the learning curves of NCKLMS1 versus a split channel approach, that treats the complex signal as two real ones using the KLMS.

The novelty criterion was used for the sparsification of the NCKLMS1  with $\delta_1=0.15$ and $\delta_2=0.2$ and of the NCKLMS2 with $\delta_1=0.1$ and $\delta_2=0.2$. In both examples, NCKLMS1 considerably outperforms the linear, widely linear (i.e., NCLMS and WL-NCLMS) and nonlinear (CNGD and MLP) algorithms (see figures \ref{FIG:equal_circu}, \ref{FIG:equal_circu2}). The NCKLMS2 also exhibits improved performance compared to the linear, widely linear and nonlinear algorithms. However, in both cases, this enhanced behavior comes at a price in computational complexity, since the NCKLMS requires the evaluation of the kernel function. In terms of the required computer time, the complexity of CKLMS1 and CKLMS2 is of the same order as the complexity of the MLP. Comparing the NCKLMS1 and the NCKLMS2, the experiments show that the results differ, with the former one leading to an improved performance. Finally, figure \ref{FIG:equal_split} illustrates that the split channel approach performs poorly compared to the NCKLMS1, especially in the circular case, as it cannot capture the correlation between the two real channels.

\subsection{Channel Identification}
\begin{figure}
\begin{center}
\includegraphics[scale=0.5]{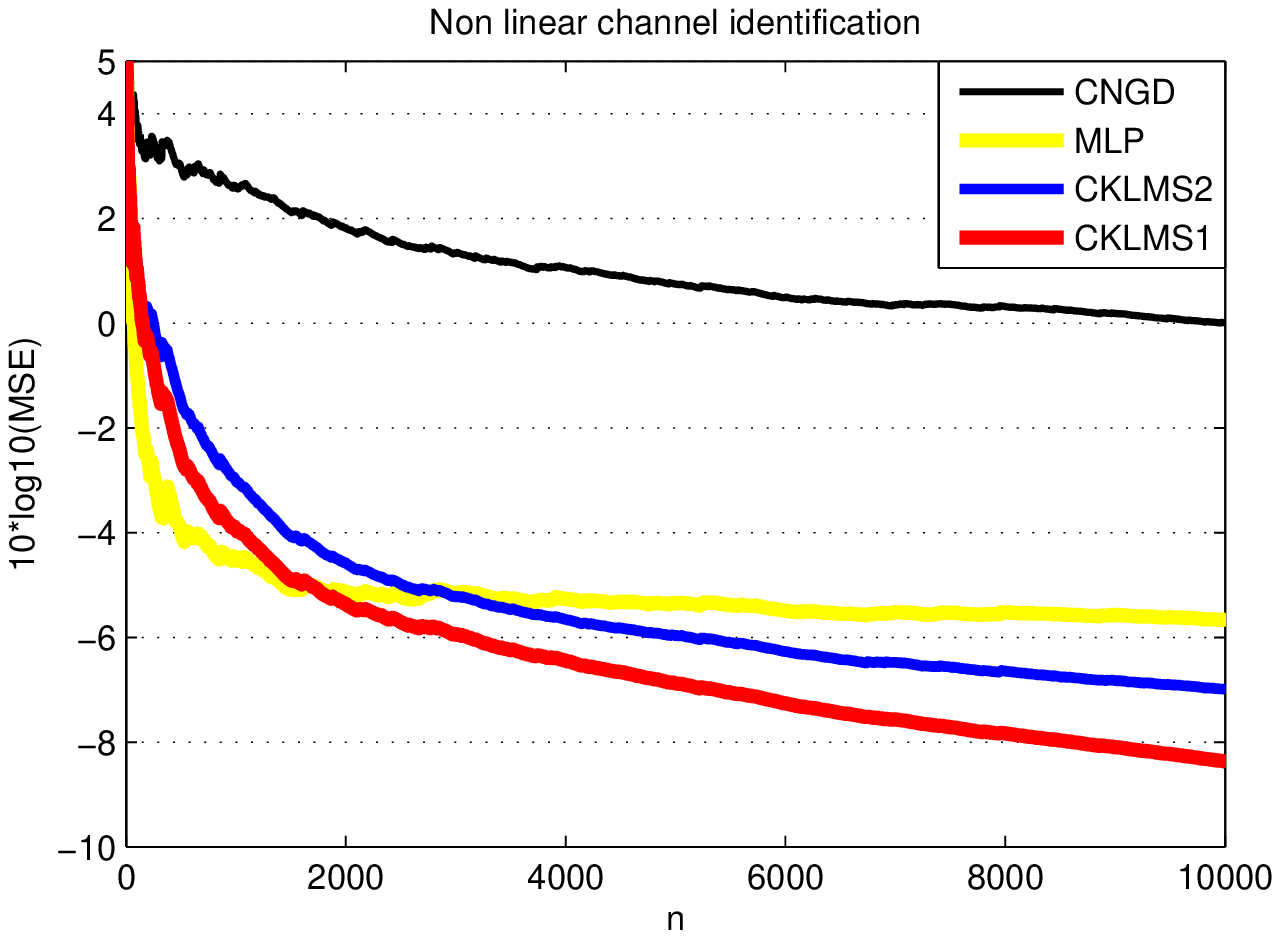}
\includegraphics[scale=0.5]{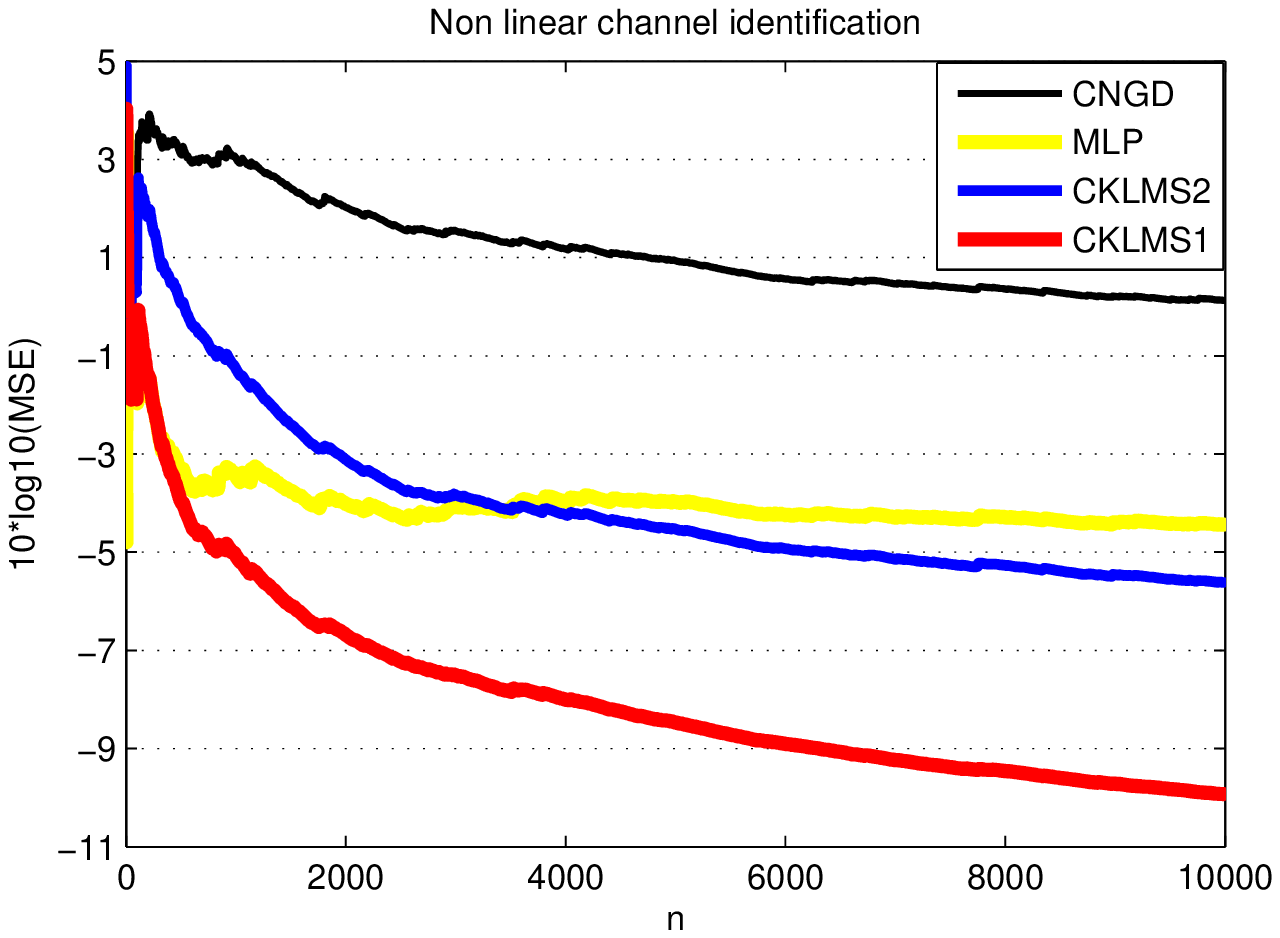}\\
(a) \hspace{18em}(b)
\end{center}
\caption{Learning curves for KÍCLMS1 ($\mu=1/2$), KÍCLMS2, ($\mu=1/4$), CNGD and $L$-50-1 MLP (filter length $L=5$) for the nonlinear channel identification problem,  for (a) the circular input case, (b) the non-circular input case ($\rho=0.1$).}\label{FIG:ident_circu}
\end{figure}

The nonlinear channel that was considered (see \cite{Adali10}) consists of a linear filter:
\begin{align*}
t(n) = \sum_{k=1}^5 h(k)\cdot s(n-k+1),
\end{align*}
where
\small{
\begin{align*}
h(k)=0.432\left(1+\cos\left(\frac{2\pi(k-3)}{5}\right)-\left(1+\cos\frac{2\pi(k-3)}{10}\right)i\right),
\end{align*}
}
for $k=1,\dots, 5$, and the nonlinear component:
\begin{align*}
x(n) = t(n) + (0.15-0.1i)t^2(n).
\end{align*}
Similar to the equalization case, the input signal that was fed to the channel had the form (\ref{EQ:input}). Experiments were conducted on a set of 10000 samples of the input signal (\ref{EQ:input}), corrupted by white gaussian noise, considering both the circular and the non-circular case. The level of the noise was set to 18dB.  Figure \ref{FIG:ident_circu}, shows the learning curves of the NCKLMS1 and the NCKLMS2  together with those obtained from the CNGD and the $L$-50-1 MLP. In this example, also, NCKLMS1 considerably outperforms both the CNGD and the $L$-50-1 MLP. The NCKLMS2 although performs better than MLP, CNGD, its performance is inferior to NCKLMS1.

%--------------------------------------------------------------------------------
\section{Conclusions}\label{SEC:Concl}
%--------------------------------------------------------------------------------
A new framework for kernel adaptive filtering for complex signal processing has been developed. The proposed methodology, besides providing a skeleton for working with pure complex kernels, allows for the construction of complex RKHSs from real ones, through a technique called complexification of RKHSs. Such an approach provides the advantage of working with popular and well understood real kernels in the complex domain. It has to be pointed out, that our method is a general one and can be used on any type of real and/or complex kernels that have or can be developed. To the best of our knowledge, this is the first time that a methodology for complex adaptive processing in RKHSs is proposed. Wirtinger's calculus has been extended to cope with the problem of differentiation in the involved (infinite) dimensional Hilbert spaces. The derived rules and properties of the extended Wirtinger's calculus on complex RKHS turn out to be similar in structure  to the special case of finite dimensional complex spaces. The proposed framework was applied on the complex LMS and two realizations for the complex Kernel LMS algorithm were developed. Experiments, which were performed on both the equalization and the identification problem of a nonlinear channel, for both circular and non-circular input data, showed a significant decrease in the steady state mean square error, compared with other known linear, widely linear and nonlinear techniques, while retaining a fast convergence.

\appendices
\section{Proof of Proposition \ref{PRO:fre_CR2}}\label{APENDIX:basic_prop}

We start with a lemma that will be used to prove the claim.
\begin{lem}\label{LEM:limit}
Consider the Hilbert space $\HH$ and $\ba, \bb\in \HH$. The limit
\begin{align}\label{EQ:lem_hilb}
\lim_{\|\bh\|_{\HH}\rightarrow 0}\frac{\langle \bh^*, \ba \rangle_{\HH} - \langle \bh, \bb\rangle_{\HH}}{\|\bh\|_{\HH}} = 0,
\end{align}
if and only if $\ba=\bb=\bZero$.
\end{lem}

Consider the first order Taylor expansions of $T_r$ and $T_i$ at $\bc=c_1 + i c_2 = (c_1,c_2)$:
\begin{align*}
T_r(\bc+\bh) &= T_r(\bc) + \left\langle h_1, \nabla_u T_r(\bc)\right\rangle_{\cH}
+ \left\langle h_2, \nabla_v T_r(\bc)\right\rangle_{\cH} + o(\|\bh\|_{\cH^2}),\\
T_i(\bc+\bh) &= T_i(\bc) + \left\langle h_1, \nabla_u T_i(\bc)\right\rangle_{\cH}
+ \left\langle h_2, \nabla_v T_i(\bc)\right\rangle_{\cH} + o(\|\bh\|_{\cH^2}).
\end{align*}
Multiplying the second relation with $i$ and adding it to the first one, we take:
\begin{align*}
\bT(\bc+\bh) &=  \bT(\bc) + \left\langle h_1,  \nabla_u T_r(\bc)  - i \nabla_u T_i(\bc)  \right\rangle_{\HH}
+ \left\langle h_2,  \nabla_v T_r(\bc)  - i \nabla_v T_i(\bc) \right\rangle_{\HH} + o(\|\bh\|_{\HH}).
\end{align*}
To simplify the notation we may define
\begin{align*}
\nabla_u \bT(\bc) = \nabla_u T_r(\bc)  + i \nabla_u T_i(\bc)\quad\nabla_v \bT(\bc) = \nabla_v T_r(\bc) + i \nabla_v T_i(\bc)
\end{align*}
and obtain:
\begin{align*}
\bT(\bc+\bh) &= \bT(\bc) + \left\langle h_1, (\nabla_u\bT(\bc))^* \right\rangle_{\HH}
  + \left\langle h_2, (\nabla_v \bT(\bc))^* \right\rangle_{\HH} + o(\|\bh\|_{\cH^2}).
\end{align*}
Next, we substitute $h_1$ and $h_2$ using the relations $h_1= \frac{\bh+\bh^*}{2}$ and $h_2=\frac{\bh-\bh^*}{2i}$ and use the sesquilinear property of the inner product of $\HH$:
\begin{align*}
\bT(\bc+\bh) &= \bT(\bc) + \frac{1}{2}\left\langle \bh, \left(\nabla_u\bT(\bc) -i\nabla_v\bT(\bc)\right)^*\right\rangle_{\HH}
 +\frac{1}{2}\left\langle \bh^*, \left(\nabla_u\bT(\bc) +i \nabla_v\bT(\bc)\right)^*\right\rangle_{\HH}  + o(\|\bh\|_{\HH}).
\end{align*}

It has already been shown  that equation (\ref{EQ:frechet_wirti2}) is essential for the development of Wirtinger's calculus. To complete the proof of the proposition we compute the fraction that appears in the definition of the complex \frechet derivative:
\begin{align*}
&\frac{\bT(\bc+\bh)-\bT(\bc) - \langle \bh, \bw\rangle_{\HH}}{\|\bh\|_{\HH}}  =\\
&\left(\frac{1}{2}\left\langle \bh, \left(\nabla_u\bT(\bc) -i\nabla_v\bT(\bc)\right)^*\right\rangle_{\HH} \right.
\left.+\frac{1}{2}\left\langle \bh^*, \left(\nabla_u\bT(\bc) + i \nabla_v\bT(\bc)\right)^*\right\rangle_{\HH}  - \langle \bh, \bw\rangle_{\HH}\right) \Big{/}\|\bh\|_{\HH}
+ \frac{o(\|\bh\|_{\HH})}{\|\bh\|_{\HH}}.
\end{align*}
Recall that, since $o(\|\bh\|_{\HH})/\|\bh\|_{\HH}\rightarrow 0$ as $\|\bh\|_{\HH}\rightarrow 0$, for this limit to exist and vanish, it is necessary that $\nabla_u\bT(\bc) + i \nabla_v\bT(\bc)=\bZero$ and $\bw^*=\nabla_u\bT(\bc) - i \nabla_v\bT(\bc)$ (see lemma \ref{LEM:limit}).
However, according to our definition,
\begin{align*}
\nabla_u\bT(\bc)  + i \nabla_v\bT(\bc)&= \left(\nabla_u T_r(\bc) - \nabla_v T_i(\bc)\right)
 + i \left(\nabla_u T_i(\bc) + \nabla_v T_r(\bc)\right).
\end{align*}
Thus, $\bT$ is differentiable in the \frechet complex sense, iff the Cauchy-Riemann conditions hold. Moreover, in this case:
\begin{align*}
\nabla\bT(\bc) =& \nabla_u T_r(\bc) + i\nabla_u T_i(\bc)
         = \nabla_v T_i(\bc) - i\nabla_v T_r(\bc).
\end{align*}

\section{Properties of Wirtinger's Derivatives on complex Hilbert spaces}\label{APPENDIX:wirti_properties}
Below we give a complete list of the main properties of the extended Wirtinger's Calculus in complex Hilbert spaces. A rigorous and detailed presentation of the theory, as well as the proofs of all these properties can be found in \cite{Bou_Wirti}.
\begin{enumerate}
\item If $\bT(\bbf)$ is $\bbf$-holomorphic at $\bc$ (i.e., it has a Taylor series expansion with respect to $\bbf$ around $\bc$), then its \frechet W-derivative at $\bc$ degenerates to the standard \frechet complex derivative and its \frechet CW-derivative vanishes, i.e., $\nabla_{\bbf^*}\bT(\bc)=\bZero$.
\item If $\bT(\bbf)$ is $\bbf^*$-holomorphic at $\bc$ (i.e., it has a Taylor series expansion with respect to $\bbf^*$ around $\bc$), then $\nabla_{\bbf}\bT(\bc)=\bZero$.
\item $\left(\nabla_{\bbf} \bT(\bc)\right)^* = \nabla_{\bbf^*} \bT^*(\bc)$.
\item $\left(\nabla_{\bbf^*} \bT(\bc)\right)^* = \nabla_{\bbf} \bT^*(\bc)$.
\item If $\bT$ is real valued, then $\left(\nabla_{\bbf} \bT(\bc)\right)^* = \nabla_{\bbf^*} \bT(\bc)$.
\item The first order Taylor expansion around $\bbf\in\HH$ is given by
\begin{align*}
\bT(\bbf+\bh) =& \bT(\bbf) + \langle \bh, \left(\nabla_{\bbf} \bT(\bbf)\right)^* \rangle_\HH
+ \langle \bh^*, \left(\nabla_{\bbf^*} \bT(\bbf)\right)^* \rangle_\HH.
\end{align*}
\item If $\bT(\bbf)=\langle \bbf, \bw\rangle_\HH$, then $\nabla_{\bbf}\bT(\bc)=\bw^*$, $\nabla_{\bbf^*}\bT(\bc)=\bZero$, for every $\bc$.
\item If $\bT(\bbf)=\langle \bw, \bbf\rangle_\HH$, then $\nabla_{\bbf}\bT(\bc)=\bZero$, $\nabla_{\bbf^*}\bT(\bc)=\bw$, for every $\bc$.
\item If $\bT(\bbf)=\langle \bbf^*, \bw\rangle_\HH$, then $\nabla_{\bbf}\bT(\bc)=\bZero$, $\nabla_{\bbf^*}\bT(\bc)=\bw^*$, for every $\bc$.
\item If $\bT(\bbf)=\langle \bw, \bbf^*\rangle_\HH$, then $\nabla_{\bbf}\bT(\bc)=\bw$, $\nabla_{\bbf^*}\bT(\bc)=\bZero$, for every $\bc$.
\item Linearity: If $\bT,\bS:\HH\rightarrow\C$ are \frechet differentiable in the real sense at $\bc\in\HH$ and $\alpha, \beta\in\C$, then
\begin{align*}
\nabla_{\bbf}(\alpha \bT + \beta \bS)(\bc) &= \alpha\nabla_{\bbf}\bT(\bc) + \beta\nabla_{\bbf}\bS(\bc)\\
\nabla_{\bbf^*}(\alpha \bT + \beta \bS)(\bc) &= \alpha\nabla_{\bbf^*}\bT(\bc) + \beta\nabla_{\bbf^*}\bS(\bc).
\end{align*}
\item Product Rule: If $\bT,\bS:\HH\rightarrow\C$ are \frechet differentiable in the real sense at $\bc\in\HH$, then:
\begin{align*}
\nabla_{\bbf} (\bT\cdot\bS)(\bc) &= \nabla_{\bbf}\bT(\bc)\bS(\bc) + \bT(\bc)\nabla_{\bbf}\bS(\bc),\\
\nabla_{\bbf^*} (\bT\cdot\bS)(\bc) &= \nabla_{\bbf^*}\bT(\bc)\bS(\bc) + \bT(\bc)\nabla_{\bbf^*}\bS(\bc).
\end{align*}
\item Division Rule: If $\bT,\bS:\HH\rightarrow\C$ are \frechet differentiable in the real sense at $\bc\in\HH$ and $\bS(\bc)\not=0$, then:
\begin{align*}
\nabla_{\bbf}\left(\frac{\bT}{\bS}\right)(\bc) &= \frac{\nabla_{\bbf}\bT(\bc) \bS(\bc) - \bT(\bc)\nabla_{\bbf}\bS(\bc)}{\bS^2(\bc)},\\
\nabla_{\bbf^*}\left(\frac{\bT}{\bS}\right)(\bc) &= \frac{\nabla_{\bbf^*}\bT(\bc) \bS(\bc) - \bT(\bc)\nabla_{\bbf^*}\bS(\bc)}{\bS^2(\bc)}.
\end{align*}
\item Chain Rule: If $\bT:\HH\rightarrow\C$ is \frechet differentiable at $\bc\in\HH$, $\bS:\C\rightarrow\C$ is differentiable in the real sense at $\bT(\bc)\in\C$,  then:
\begin{align*}
\nabla_{\bbf}\bS\circ\bT(\bc) &= \frac{\partial \bS}{\partial z}(\bT(c))\nabla_{\bbf}\bT(\bc)
 + \frac{\partial \bS}{\partial z^*}(\bT(\bc))\nabla_{\bbf}(\bT^*)(\bc),\\
\nabla_{\bbf^*}\bS\circ\bT(\bc) &= \frac{\partial \bS}{\partial z}(\bT(\bc))\nabla_{\bbf^*}\bT(\bc)
 + \frac{\partial \bS}{\partial z^*}(\bT(\bc))\nabla_{\bbf^*}(\bT^*)(\bc).
\end{align*}
\end{enumerate}

The proofs of properties 1 and 2 are rather obvious. Here, we give the proofs of properties 3, 7 and 11, which have been used to derive the main results of this paper.
\begin{proof}[Proof of property 3]
The existence of $\nabla_{\bbf}\bT(\bc)$ and $\nabla_{\bbf^*}\bT(\bc)$ is guaranteed by the \frechet differentiability of $\bT$ at $\bc$ (in the real sense). To take the result, observe that:
\begin{align*}
\left(\nabla_{\bbf}\bT(\bc)\right)^* =&  \frac{1}{2}\left(\nabla_u T_r(\bc) + \nabla_v T_i(\bc)\right)
      - \frac{i}{2}\left(\nabla_u T_i(\bc) - \nabla_v T_r(\bc)\right)\\
=&  \frac{1}{2}\left(\nabla_u T_r(\bc) - \nabla_v(-T_i)(\bc)\right)
      + \frac{i}{2}\left(\nabla_u(-T_i)(\bc) + \nabla_v T_r(\bc)\right)\\
=&\left(\nabla_{\bbf^*}\bT^*(\bc)\right).
\end{align*}
\end{proof}
Property 4 can be proved similarly.

\begin{proof}[Proof of property 7]
Considering the definition of \frechet complex derivative (see equation \ref{EQ:frechet_complex}), we observe that:
\begin{align*}
\bT(\bc + \bh) - \bT(\bc) - &\langle\bh, \bg\rangle_{\HH} =
  \langle \bc+\bh, \bw\rangle_{\HH} - \langle\bc,\bw\rangle_{\HH} - \langle\bh, \bg\rangle_{\HH}
 =\langle \bh, \bw\rangle_{\HH} - \langle \bh, \bg\rangle_{\HH}.
\end{align*}
Thus, $\bT$ is \frechet complex differentiable at $\bc$, with $\nabla\bT(\bc)=\bw^*$ and from property 1, $\nabla_{\bbf^*}(\bc)=\bZero$ and $\nabla_{\bbf}(\bc)=\bw$.
\end{proof}

\begin{proof}[Proof of property 11]
Let $\bT(\bbf) = T_r(u_{\bbf},v_{\bbf}) + i T_i(u_{\bbf},v_{\bbf})$, $\bS(\bbf) = \bS(u_{\bbf}+i v_{\bbf}) = S_r(u_{\bbf},v_{\bbf}) + i S_i(u_{\bbf},v_{\bbf})$ be two complex functions and $\alpha, \beta\in\C$, such that $\alpha=\alpha_1 + i\alpha_2$, $\beta=\beta_1 + i\beta_2$. Then $\bR(\bbf) = \alpha \bT(\bbf) + \beta \bS(\bbf)$
and the \frechet W-derivative of $\bR$ will be given by:
\begin{align*}
\nabla_{\bbf}\bR(\bc) =& \frac{1}{2}\left(\nabla_u R_r(\bc) + \nabla_v R_i(\bc)\right)
      + \frac{i}{2}\left( \nabla_u R_i(\bc) -  \nabla_v R_r(\bc)\right).
\end{align*}
Applying the linearity property of the ordinary \frechet derivative, after some algebra we take the result.
For the second part, in view of properties 3, 4 and the linearity property of the \frechet W-derivative, the \frechet CW-derivative of $\bR$ at $\bc$ will be given by:
\begin{align*}
\nabla_{\bbf^*}\bR(\bc) =& \nabla_{\bbf^*}(\alpha \bT+\beta \bS)(\bc)
 = \left(\nabla_{\bbf}(\alpha \bT+\beta \bS)^*(\bc)\right)^*\\
    =& \left(\nabla_{\bbf}(\alpha^*\bT^*+\beta^*\bS^*)(\bc)\right)^*
    = \left(\alpha^*\nabla_{\bbf}\bT^*(\bc) + \beta^*\nabla_{\bbf}\bS^*(\bc) \right)^*\\
    =& \alpha \left(\nabla_{\bbf}\bT^*(\bc)\right)^* + \beta \left(\nabla_{\bbf}\bS^*(\bc)\right)^*
    = \alpha \nabla_{\bbf^*}\bT(\bc) + \beta \nabla_{\bbf^*}\bS(\bc),
\end{align*}
which completes the proof.
\end{proof}

\ifCLASSOPTIONcaptionsoff
  \newpage
\fi



\bibliographystyle{IEEEtran}
\bibliography{refs}

%\begin{biography}[{\includegraphics[width=1in,height=1.25in,clip,keepaspectratio]{mshell}}]{Michael Shell}

\begin{IEEEbiography}{Pantelis Bouboulis} (M' 10)
received the M.Sc. and Ph.D. 	
degrees in informatics and telecommunications from 	
the National and Kapodistrian University of Athens, 	
Greece, in 2002 and 2006, respectively. 	
From 2007 till 2008, he served as an Assistant Professor in the Department of Informatics and Telecommunications, University of Athens. His current research interests lie in the areas of machine learning, 	
fractals, wavelets and image processing.
He is a member of AMS and IEEE.
\end{IEEEbiography}

\begin{IEEEbiography}{Sergios Theodoridis}
(F'08) is currently Professor of
signal processing and communications in the Department
of Informatics and Telecommunications, University
of Athens, Athens, Greece. His research interests
lie in the areas of adaptive algorithms and communications,
machine learning and pattern recognition,
and signal processing for audio processing and
retrieval. He is the co-editor of the book Efficient Algorithms
for Signal Processing and System Identification
(Prentice-Hall, 1993), coauthor of the best
selling book Pattern Recognition (Academic, 4th ed.,
2008), and the coauthor of three books in Greek, two of them for the Greek Open
University.
He is currently an Associate Editor of the IEEE TRANSACTIONS ON NEURAL
NETWORKS, the IEEE TRANSACTIONS ON CIRCUITS AND SYSTEMS II, and a
member of the editorial board of the EURASIP Wireless Communications
and Networking. He has served in the past as an Associate Editor of the IEEE
TRANSACTIONS ON SIGNAL PROCESSING, the IEEE Signal Processing Magazine,
the EURASIP Journal on Signal Processing, and the EURASIP Journal
on Advances on Signal Processing. He was the general chairman of EUSIPCO
1998, the Technical Program cochair for ISCAS 2006, and cochairman of
ICIP 2008. He has served as President of the European Association for Signal
Processing (EURASIP) and he is currently a member of the Board of Governors
for the IEEE CAS Society. He is the coauthor of four papers that have received
best paper awards including the 2009 IEEE Computational Intelligence Society
Transactions on Neural Networks Outstanding paper Award. He serves as an
IEEE Signal Processing Society Distinguished Lecturer. He is a member of
the Greek National Council for Research and Technology and Chairman of the
SP advisory committee for the Edinburgh Research Partnership (ERP). He has
served as vice chairman of the Greek Pedagogical Institute and for four years,
he was a member of the Board of Directors of COSMOTE (the Greek mobile
phone operating company). He is Fellow of IET and a Corresponding Fellow
of FRSE.
\end{IEEEbiography}

%\newpage


%\vfill


\end{document}